\documentclass[lettersize,journal]{IEEEtran}
\usepackage{amsmath,amsfonts}
\usepackage{algorithmic}
\usepackage{algorithm}
\usepackage{array}
\usepackage[caption=false,font=normalsize,labelfont=sf,textfont=sf]{subfig}
\usepackage{textcomp}
\usepackage{stfloats}
\usepackage{url}
\usepackage{verbatim}
\usepackage{graphicx}
\usepackage{cite}
\usepackage[breaklinks]{hyperref}
\hypersetup{hidelinks}
\usepackage{makecell}
\usepackage{bbding}
\usepackage{pifont}
\usepackage{wasysym}
\usepackage{amssymb}
\usepackage{gensymb}
\usepackage{dsfont}
\usepackage{multirow}
\usepackage{threeparttable}
\usepackage{bm}
\usepackage{balance}
\usepackage{graphicx}
\usepackage{xcolor}
\usepackage{float}
\usepackage{booktabs}
\newcommand{\revise}[1]{{\color{black}{#1}}}

\begin{document}

\title{Efficient Alignment of Unconditioned Action Prior for Language-conditioned Pick and Place in Clutter}

\author{Kechun Xu, Xunlong Xia, Kaixuan Wang, Yifei Yang, Yunxuan Mao, \\ Bing Deng, Jieping Ye, Rong Xiong, Yue Wang
        % <-this % stops a space
% \thanks{This paper was produced by the IEEE Publication Technology Group. They are in Piscataway, NJ.}% <-this % stops a space
% \thanks{Manuscript received April 19, 2021; revised August 16, 2021.}
\thanks{This work was supported by the Joint Funds of the National Natural Science Foundation of China under Grant U24A20128 and the Zhejiang Provincial Natural Science Foundation of China under Grant LD25F030001.}
\thanks{Kechun Xu is with Zhejiang University and Alibaba Cloud, Hangzhou, China. Xunlong Xia, Bing Deng, and Jieping Ye are with Alibaba Cloud, Hangzhou, China. Kaixuan Wang, Yifei Yang, Yunxuan Mao, Rong Xiong, and Yue Wang are with Zhejiang University, Hangzhou, China. Corresponding author, {\tt\small wangyue@iipc.zju.edu.cn}. }
% \thanks{Project page: {\href{https://xukechun.github.io/papers/A2/}{https://xukechun.github.io/papers/A2}}.
% }
}

% The paper headers
\markboth{Journal of \LaTeX\ Class Files,~Vol.~14, No.~8, August~2021}%
{Shell \MakeLowercase{\textit{et al.}}: A Sample Article Using IEEEtran.cls for IEEE Journals}

% \IEEEpubid{0000--0000/00\$00.00~\copyright~2021 IEEE}
% Remember, if you use this you must call \IEEEpubidadjcol in the second
% column for its text to clear the IEEEpubid mark.

\maketitle

\begin{abstract}

We study the task of language-conditioned pick and place in clutter, where a robot should grasp a target object in open clutter and move it to a specified place. Some approaches learn end-to-end policies with features from vision foundation models, requiring large datasets. Others combine foundation models in a zero-shot setting, suffering from cascading errors. In addition, they primarily leverage vision and language foundation models, focusing less on action priors. In this paper, we aim to develop an effective policy by integrating foundation priors from vision, language, and action. We propose A$^2$, an action prior alignment method that aligns unconditioned action priors with 3D vision-language priors by learning one attention layer. The alignment formulation enables our policy to train with less data and preserve zero-shot generalization capabilities. We show that a shared policy for both pick and place actions enhances the performance for each task, and introduce a policy adaptation scheme to accommodate the multi-modal nature of actions. Extensive experiments in simulation and the real-world show that our policy achieves higher task success rates with fewer steps for both pick and place tasks in clutter, effectively generalizing to unseen objects and language instructions. 
Videos and codes are available at {\href{https://xukechun.github.io/papers/A2/}{https://xukechun.github.io/papers/A2}}.

\end{abstract}

 \def\abstractname{Note to Practitioners}

\begin{abstract}
This research is motivated by the challenge of generalizable policy learning of language-conditioned pick and place in clutter. Solving such a challenge could significantly improve the robot's level of automation and intelligence in household and industrial pick and place tasks. Existing methods struggle with large data requirements, poor generalization to unseen scenarios, and cascading errors across individual components. To overcome these limitations, we propose to integrate priors from vision, language, and action foundation models by learning-based alignment. Our policy aligns action priors with 3D vision-language priors by learning one attention layer, requiring less data and preserving zero-shot generalization capabilities from foundation models. Experiments show that our method can improve both task success rate and generalization for pick and place tasks in simulation and the real world. In future work, we will incorporate more action foundation models to extend our approach of action prior alignment to a wider range of tasks, offering a promising direction for general manipulation. 

\end{abstract}

\section{Introduction}
\label{sec:intro}

\IEEEPARstart{T}{he} ability to pick and place objects is essential for robotic manipulation~\cite{xu2021efficient,zeng2021transporter,qureshi2021nerp,goyal2022ifor,tang2023selective,xu2024grasp}. Consider a scenario where a robot is commanded with language instructions to grasp a target object in open clutter, and move it to a specified place. The target object may be partially or fully occluded, posing challenges for object grounding and grasping. In such scenarios, multiple pick and place actions may be needed to clear obstacles for object rearrangement. 

A common way to construct a policy for such tasks is to predict 6-DoF actions directly from raw sensory information, as in classic end-to-end policies. Recently, these policies have achieved promising performances by incorporating features of pre-trained foundation models, {\it e.g.}, vision-language models~(VLM) and large language models~(LLM)~\cite{shridhar2022cliport,shridhar2023perceiver,jiang2023vima,ze2023gnfactor,gervet2023act3d,goyal2024rvt2}. However, they require large amounts of demonstration data for policy learning, particularly for tasks involving cluttered environments. In addition, one has to deal with generalization issues to deploy these policies in real-world applications.

\begin{figure}[t]
  \centering
  % \color{blue}
  \includegraphics[width=\linewidth]{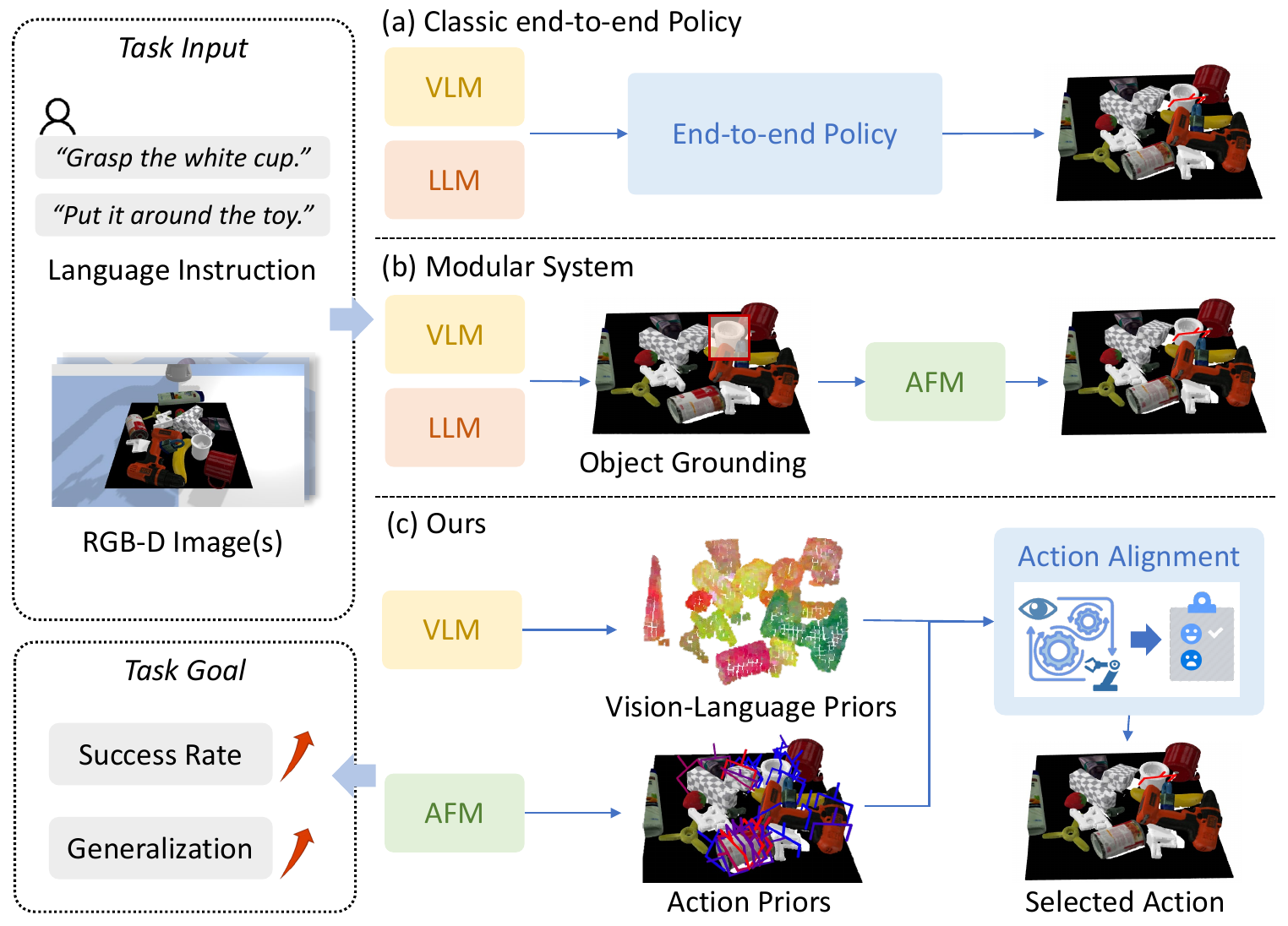}
  \vspace{-0.6cm}
  \caption{
  Compared to previous methods (a) classic end-to-end policies and (b) modular systems, our method integrates foundation priors from vision, language, and action through alignment by one attention layer, which enables more efficient policy learning and better task performance. 
  }
  \label{fig:teaser}
  \vspace{-0.6cm}
\end{figure}

In contrast, other methods harness the zero-shot generalization capabilities of foundation models by developing modular systems. Many works investigate visual representations for object grounding, followed by rule-based action planners for object manipulation~\cite{goodwin2022semantically,ji2024graspsplats,shorinwa2024splat,zheng2024gaussiangrasper,jatavallabhula2023conceptfusion,wang2024d}. For example, LERF-TOGO~\cite{rashid2023lerftogo} builds 3D scene representations by distilling features from vision-language models, then performs object grounding to filter candidate actions generated by an action foundation model. These approaches are mostly learning-free, showcase zero-shot generalization, and utilize action candidates as priors. Nevertheless, they demand high accuracy in visual grounding, which remains challenging in cluttered settings. Even with correct grounding, the target in clutter may be ungraspable.
Some works employ large language models as planners to decide the object grasp order in clutter, but still suffer from cascading errors across individual modules~\cite{Ahn2022DoAI,huang2023voxposer,qian2024thinkgrasp}.

In general, end-to-end methods require large datasets to effectively learn a policy with substantial network parameters and pay less attention to action priors, whereas modular systems struggle with cascading errors when combining several foundation models in a zero-shot setting. 
Considering that action foundation models can provide action priors that are unconditioned on specific tasks, we raise a question: \textit{Given unconditioned action priors, is there a policy that can improve performance while learning fewer network parameters?}

To leverage unconditioned action priors in specific tasks, we adopt the idea of \textit{alignment} with a reward model, inspired by the RLHF technique in large language model training~\cite{bai2022training}. Taking action foundation models as generators, we build a probabilistic policy upon the generated actions for reward alignment. For specific pick and place tasks, the reward model can be defined as a simple binary function. Then, expert demonstrations can be extended into state-action pairs with binary scores. In this way, we can learn the policy that aligns with the task reward by maximizing the probabilities of demonstrated actions through imitation learning.

Guided by the insights, we propose A$^2$, an \textbf{A}ction Prior \textbf{A}lignment method that aligns unconditioned action priors based on task-conditioned vision-language priors by \textit{learning one attention layer}~(Figure~\ref{fig:teaser}). Action foundation models, such as GraspNet~\cite{fang2020graspnet}, generate action candidates, providing unconditioned action priors and largely reducing the action space. For vision and language input, we construct 3D zero-shot representations combining vision-language foundation priors from the vision-language model MaskCLIP~\cite{zhou2022extract}. 
Based on these priors, we perform alignment by a cross-attention layer to predict action probabilities for planning. In this way, our policy is to learn one-dimensional probabilities over action priors, requiring less training data and preserving zero-shot generalization capabilities. To learn such a policy, we construct a score-based dataset from expert demonstrations. We use shared network parameters for pick and place tasks, improving performance simultaneously for each task. We also propose a fast policy adaptation scheme, allowing fine-tuning for action multi-modality modeling. At inference time, our policy aligns actions across the scene to predict a sequence of grasps to remove obstacles for target grasping, and ultimately place the target at the specified location. A wide range of experiments in both simulation and real-world settings show that our policy achieves higher task success rates with fewer planning steps, with zero-shot generalization to unseen objects and language instructions. To summarize, our main contributions are:

\begin{itemize}
    \item We propose A$^2$, an efficient action prior alignment method that allows learning one attention layer for language-conditioned pick and place in clutter. 
    \item We leverage the vision-language model to construct 3D vision-language priors that indicate task information with zero-shot generalization capability.
    \item We conduct alignment of unconditioned action priors based on vision-language priors from foundation models.
    \item We propose to use shared parameters for pick and place, and develop a fast policy adaptation mechanism for action multi-modality modeling.
    \item The learned policy is evaluated on a series of scenarios with seen and unseen objects and language instructions in both simulated and real-world settings, of which the results validate the effectiveness and generalization.
\end{itemize}

\section{Related Works}
\label{sec:related}

\subsection{Target-oriented Pick and Place in Clutter}
Robotic pick and place in clutter has been a topic of interest in manipulation for decades. Traditional approaches~\cite{king2016rearrangement,xu2022efficient,tian2022sampling,xu2023failure} are in the context of task and motion planning~(TAMP) under the assumption of known object models and states. These methods struggle in open real scenarios, where obtaining precise object models and states is challenging. More recent research studies target-oriented unknown object grasping in clutter by first clearing obstacles~\cite{zeng2022robotic, murali20206, fang2018multi}, or retrieving the target object through non-prehensile actions~\cite{danielczuk2018linear,kurenkov2020visuomotor, yang2020deep, xu2021efficient}. \cite{zeng2021transporter,goyal2022ifor,xu2024grasp}\revise{\cite{zhai2023monograspnet,zhai2024sg}} step forward to build unknown object pick and place systems in cluttered environments, which are promising for real applications. However, these works still require images to specify target objects. Instead, language instructions are more flexible in open-world applications. By cooperating with foundation models, policies are capable of dealing with open-vocabulary objects in scattered scenes~\cite{shridhar2022cliport,vuong2023grasp,jiang2023vima,jia2024open}. In this paper, we aim to develop a policy for open-vocabulary pick and place in clutter, with the target specified with language instructions.

\subsection{\!\!Foundation Models for Language-conditioned Manipulation}
Foundation models in the field of CV and NLP have demonstrated powerful performance~\cite{radford2021learning,kirillov2023segment,oquab2023dinov2,2023GPT4VisionSC}, and have been explored to facilitate robotic manipulation in open-world applications. A common way to utilize foundation models is to directly ground their capabilities into robotic scenarios. A series of approaches~\cite{goodwin2022semantically,Ahn2022DoAI,huang2023instruct2act} uses vision foundation models for object grounding from flexible language instructions. Among them, some works explore object-centric representations for better scene understanding~\cite{goodwin2022semantically,xu2023joint,zhu2023viola,yang2024attribute,yuanm2t2}. Other methods build 3D scene representations capturing both semantic and geometric information~\cite{shim2023snerl,jiang2021synergies,simeonov2022neural,shafiullah2022clip,ke20243d,deng2024openobj}. For example, several approaches distill 3D neural feature fields from 2D foundation models~\cite{shen2023distilled,rashid2023lerftogo}, requiring dense camera views and time-consuming training for high-quality rendering. This hinders real-time interaction in real-world scenarios. And efforts to overcome these limitations include introducing 3D Gaussian Splatting~\cite{zheng2024gaussiangrasper,ji2024graspsplats,shorinwa2024splat} and using sparse-view 3D representations~\cite{wang2024d,jatavallabhula2023conceptfusion}. There are also methods~\cite{liang2023code,huang2023voxposer,vemprala2024chatgpt,qian2024thinkgrasp,wake2024gpt,hu2024look} utilizing the reasoning capability of large language models to build systems for planning. However, the performance of these policies largely depends on the capability of foundation models, and suffers from cascaded errors across individual modules. Another line of works~\cite{shridhar2022cliport,shridhar2023perceiver,ze2023gnfactor,jiang2023vima,gervet2023act3d,goyal2024rvt2,ke20243d} integrates features from vision foundation models into end-to-end policies. Despite promising results, these works consume extensive demonstration data and take plenty of training steps for convergence. In addition, one has to face the generalization issue if the tested objects or scenes are significantly different from those in the training data.

Recently, researchers have tried to learn action foundation models from large-scale robot data. For instance, AnyGrasp~\cite{fang2023anygrasp} is a grasp foundation model capable of generating grasp actions for open scenes. More generally, efforts are made to develop large Vision-Language-Action models~(VLA) for general tasks and even embodiments~\cite{o2023open,team2024octo,kim2024openvla,liu2024rdt,black2024pi_0}. However, leveraging priors from these action foundation models is much less explored. Some methods deploy pre-trained grasp models to generate grasp actions after object grounding, essentially paying less attention to action planning~\cite{rashid2023lerftogo,yenamandra2023homerobot}. In this paper, our policy aims to integrate priors from vision, language, and action foundation models to improve task performance.

\begin{figure*}[t]
  \centering
  \includegraphics[width=\textwidth]{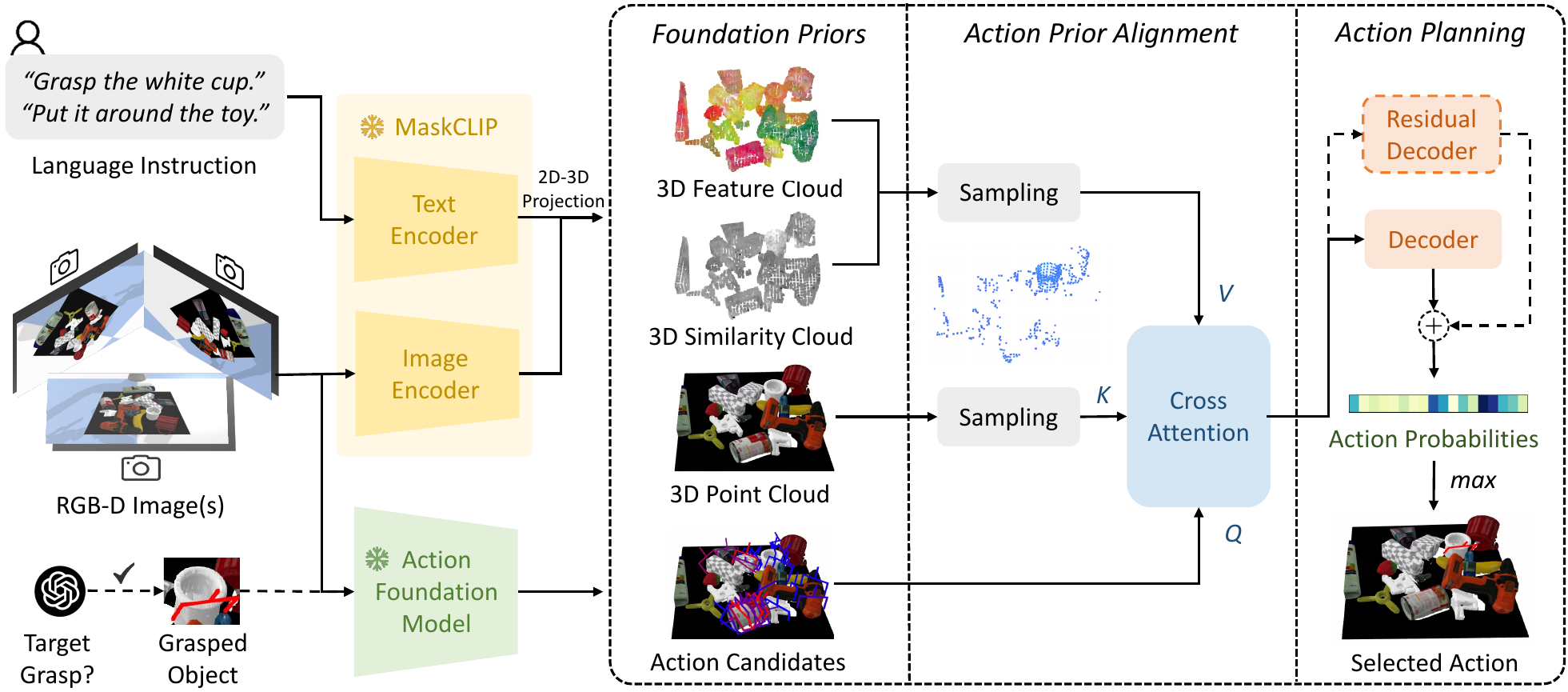}
  \vspace{-0.5cm}
  \caption{
  \revise{{\bf Overview}. Given the language instruction and RGB-D image(s), the vision-language model MaskCLIP~\cite{zhou2022extract} extracts dense patch-level features, which are projected into 3D representations, including a feature cloud, a similarity cloud, and a point cloud. In addition, the action foundation model generates action candidates. Based on these foundation priors, our policy conducts alignment for action planning.}}
  \vspace{-0.5cm}
  \label{fig:overview}
\end{figure*}

% \section{Method}
\label{sec:method}

%-------------------------------------------------------------------------
\section{Overview}
{\bf Unconditioned Action Priors based Policy.} Given the RGB-D image(s) $\mathcal{I}\!=\!\{I_i\}_{i=0,1,...,M}$ and the language instruction $\mathcal{L}$, we leverage foundation models to extract vision, language, and action priors. Consider an action foundation model that generates $L$ candidate actions from image(s) as action priors $\mathcal{A}_L(\mathcal{I})=\{a_k\}_{k=0,1,...,L}$, $L$ generally has a controllable upper limit. These priors, distilled from a wide range of unconditioned data, provide feasible action patterns for downstream tasks and largely narrow the action space. Upon these priors, we construct a probabilistic policy $\pi$.
\begin{equation}
\begin{aligned}
\pi\left(a|\mathcal{I},\mathcal{L}\right)&=\sum_{k=1}^L \omega\left(a_k|\mathcal{I},\mathcal{L}\right) \delta\left(a-a_k\right) \\
\mathrm{s.t.}\quad&\sum_{k=1}^L \omega\left(a_k|\mathcal{I},\mathcal{L}\right)=1
\end{aligned}
\end{equation}
where $\omega\left(a_k|\mathcal{I},\mathcal{L}\right)$ demonstrates the probability of $a_k$ conditioned on the vision and language information.

{\bf Alignment with Reward.} Modular systems obtain $\omega$ with rule-based filtering upon visual grounding results, which demands high visual accuracy. Instead, we propose to learn the $\omega$ to align unconditioned action priors based on vision-language priors. In this way, our policy is to learn one-dimensional probabilities over action priors, largely alleviating data demands. Consider this alignment problem via RL objective, let $r(a,\mathcal{I},\mathcal{L})$ denote the reward function, then the optimal policy is to maximize the expected sum of future rewards. For pick and place tasks, $r(a,\mathcal{I},\mathcal{L})$ can be easily defined as 
\begin{equation}
r(a,\mathcal{I},\mathcal{L}) = \begin{cases}1, & \text{pick or place successfully} \\
0, & \text{otherwise} \end{cases} \\
\end{equation}
{\bf Alignment by Imitation Learning.} By employing expert planners, we can collect demonstrations $\mathcal{D}=\{\mathcal{I}_d,\mathcal{L}_d, a_d\}$, where $a_d\in \mathcal{A}_L(\mathcal{I}_d)$. Then we have $r(a_d,\mathcal{I}_d,\mathcal{L}_d)=1$. Therefore, we can augment each demonstration into score-based samples by labeling $a_d$ as 1, with the remaining ones in $\mathcal{A}_d$ as 0. In this way, we can learn the policy $\pi$ that aligns with the reward by maximizing the likelihood of $a_d$ for $\mathcal{I}_d,\mathcal{L}_d$ through imitation learning.
\revise{\begin{equation}
\label{eq:imitation}
\max_{a_d \in \mathcal{A}_L(\mathcal{I}_d)} \omega\left(a_d|\mathcal{I}_d,\mathcal{L}_d\right)
\end{equation}}

{\bf Architecture of A$^2$.} Fig.~\ref{fig:overview} presents the pipeline of our method. For vision-language input, our method extracts dense patch-level features using MaskCLIP~\cite{zhou2022extract}. Then the features are projected into 3D representations, including a 3D point cloud, a 3D feature cloud, and a 3D similarity cloud. Specifically, each coordinate in the point cloud corresponds to a visual feature and a task-relevant vision-language similarity. Additionally, we utilize the action foundation model to yield a set of action candidates. Based on the vision, language, and action priors, we propose to conduct action prior alignment for action planning. We first sample points with higher similarity to create a more compact representation. Then a cross-attention transformer takes action features as queries, 3D position features as keys, and 3D vision-language features as values to align action priors conditioned on vision-language information. The output fusion features are fed into a decoder to get the probabilities of candidate actions. 

As a system, our policy first receives the language instruction to grasp the target object, and predicts a sequence of grasp actions by closed-loop action alignment. 
\revise{If the target object is not grasped, our policy will remove the grasped obstacles and proceed regrasping. Once the target is successfully grasped, our policy takes the language instruction of placement along with the grasped object for place prediction, finally placing the grasped object in the assigned location.}

\section{Foundation Priors}
\subsection{3D Vision-Language Priors}
We leverage the zero-shot generalization capability of foundation models to construct 3D visual representations that convey semantic and task-relevant information, which can be updated in real-time.

{\bf Generalizable Visual-Language Features.} We extract features through the pre-trained vision-language model CLIP~\cite{radford2021learning} that maps visual and
 language embeddings by training on millions of image-text data. However, CLIP originally generates image-level features. To obtain denser features, we apply MaskCLIP~\cite{zhou2022extract} reparameterization trick to extract patch-level features from CLIP. To further get more fine-grained features, we crop each RGB image into several sub-images to extract patch-level features, and concatenate them together to form the final visual feature map.

{\bf 3D Representations.} Given RGB-D image(s) $\mathcal{I}\!=\!\{I_i\}_{i=0,1,...,M}$ from one or more cameras with fixed viewpoints, we first extract a 3D point cloud $\mathbf{p}$ within the workspace using the camera parameters. For each point $p_j$ of $\mathbf{p}$, we project it back to $i$th camera viewpoint as the pixel $u_j^i$, and get its visual feature $f_j^i$ by interpolation. Following \cite{wang2024d}, we compute weights for each camera according to the visibility and distance of $p_j$ in the corresponding camera. Finally, we fuse features from all camera viewpoints using a weighted sum, denoted as $f_j$. More details can be accessed in Appendix. 

{\bf 3D Feature Cloud.} Each point $p_j$ within the workspace paired with its feature $f_j$, forms the 3D feature cloud $\mathbf{f}$. This representation implies the visual information of the scene, which is semantic and zero-shot generalizable. 

{\bf 3D Similarity Cloud.} To represent the task-relevant information, we further utilize the vision-language similarity property of MaskCLIP. Specifically, the language instruction is encoded by the MaskCLIP text encoder. For each point $p_j$, we compute the cosine similarity between the language embedding and the visual feature $f_j$ to get a similarity value $s_j$, resulting in a 3D similarity cloud $\mathbf{s}$. This representation reflects the degree of task relevance of each point.

\subsection{Unconditioned Action Priors}

{\bf Action Foundation Models.} We employ different action foundation models to yield candidate actions for pick and place respectively. For object picking, we adopt the pre-trained GraspNet~\cite{fang2020graspnet} to generate 6-DoF grasp poses that demonstrate feasible grasp actions for all objects across the whole scene. For object placement, we first obtain all the object region proposals, then place poses are sampled in and around each object region without overlapping with each other. 

{\bf Action Candidates.} By utilizing action foundation models, we yield a set of $L$ candidate actions $\mathcal{A}_L(\mathcal{I})=\{a_k\}_{k=0,1,...,L}$. $L$ generally has a controllable upper limit and is variable in different scenarios. These candidate actions provide unconditioned priors of the way to manipulate objects and largely narrow the action space into a limited set, facilitating efficient policy learning.

\section{Action Prior Alignment}
Based on foundation priors, we propose to conduct action prior alignment by learning one attention layer. 

% {\bf Policy Architecture.} 
\subsection{Alignment Architecture}
Considering that directly taking complete 3D representations is sample-inefficient, we first conduct prioritized sampling to get a more compact representation. To be specific, we sample $N$ points with higher similarities to generate sampled 3D representations $\mathbf{p}_N$, $\mathbf{f}_N$ and $\mathbf{s}_N$. Note that $N$ is a hyperparameter closely related to the total number of 3D points in the representations. Empirically, we sample half of the points from the workspace. Given the sampled 3D visual representations and the generated action candidates, we perform action prior alignment via cross-attention to obtain fusion features, followed by a decoder to predict the action probabilities. Finally, the action with the highest probability is selected for execution. 

\subsection{Cross Attention}
We propose to align unconditioned action priors based on task-conditioned vision-language priors. To be specific, we employ transformer’s attention mechanism~\cite{vaswani2017attention}: $\text{Attention}(Q, K, V)=\text{Softmax}\left({Q K^T}\right) V$, where $Q, K, V$ denote query, key and value respectively. 

We weight the 3D visual features $\mathbf{f}_N$ with the similarity values $\mathbf{s}_N$, which capture the vision-language information. We encode $L$ action pose features by an MLP to generate action features. The 3D points $\mathbf{p}_N$ are projected into a nonlinear space using positional embedding as in \cite{mildenhall2020nerf}, followed by an MLP to encode position features. To align action features based on vision-language information, the cross-attention transformer takes $L$ action pose features as queries, $N$ position features as keys, and $N$ vision-language features as values, outputting $L$ fusion features $\mathcal{F}_L$. We use RoPE~\cite{su2024roformer} to encode relative position embeddings for keys and values.
\begin{equation}
\begin{aligned}
Q&=\mathrm{MLP_1}\left(\mathcal{A}_L\right) \\
K&=\mathrm{RoPE}\left(\mathrm{MLP_2}\left(\mathbf{p}_N\right)\right) \\
V&=\mathrm{RoPE}\left(\mathbf{f}_N\circ\mathbf{s}_N\right) \\
\end{aligned}
\end{equation}

\subsection{Policy Learning}
{\bf Shared Policy for Pick and Place.} We train the policy with demonstration data collected by model-based expert planners, and propose to train a policy for pick and place with shared parameters. That is, after generating the 3D representations and candidate actions, pick and place share the same information for action alignment. This is because there is strong common information between pick and place actions. In cluttered scenes, both pick and place tasks require the policy to focus on the regions close to the target. To pick a target object in clutter, the robot should first move away the obstacles hindering the grasping of target object, and most of the time the obstacles locate close to the target. For placement, there is an additional requirement to distinguish spatial relations, but focusing around the reference object still helps.

{\bf Policy Adaptation for Multi-modality Modeling.} In fact, for both pick and place tasks, the distribution of actions is inherently multi-modal. In particular, for place tasks, the multi-modal characteristic is more significant, {\it e.g.} when placing around an object, there may be several feasible actions. However, due to the difficulty of executing all actions in each step, the demonstration data labels only one action as the ground truth. This potentially misleads the policy and degenerates the multi-modality modeling of actions. To address this issue, we propose a policy adaptation scheme using a residual block:
\begin{equation}
\begin{aligned}
\Omega_L&=\mathrm{Decoder}\left(\mathcal{F}_L\right) \\
\Omega_L^{r}&=\mathrm{Decoder}^{r}\left(\mathcal{F}_L\right) \\
\Omega_L^\prime&=\alpha \Omega_L + (1-\alpha) \Omega_L^{r} \\
\end{aligned}
\end{equation}
where $\Omega_L=\{\omega_k\}_{k=1}^L$ represents the original predicted action probabilities of $\mathcal{A}_L$, $\Omega_L^{r}$ is the residual output of probabilities, and $\Omega_L^\prime$ is the weighted sum of $\Omega_L$ and $\Omega_L^{r}$. By fine-tuning the policy with a small set of multi-labeled data of place tasks, we can further improve the policy performance~(Sec.~\ref{sec:ablation}).

{\bf Flexible Language Instructions.} Our policy is able to deal with flexible language instructions without assigning concrete object labels. 
\revise{For pick tasks, it can handle language instructions like ``Give me the \{target\}'' or ``Get something to \{target\}'', where \{target\} can be a concrete label~({\it e.g. }banana), a general category~({\it e.g. }fruit), or the attribute of color~(e.g. red), shape~(e.g. round), or even a functional description~({\it e.g. }hold other things).} 
For place tasks, the language instructions are similar, but with additional spatial relation words, such as ``Move the object \{relation\} the \{reference\}''. Here \{relation\} specifies the spatial relationship respective to the \{reference\}. \{reference\} is analogous to \{target\}, while \{relation\} can be words indicating ``on'' or ``around'' relations relative to {reference\}. For instance, words like ``on top of'', and ``into'' belong to ``on'' relation, and others such as ``next to'', and ``near'' belong to ``around'' relation.

\section{Implementation Details}
\label{sec:method-details}
{\bf Simulation Environment.} We collect demonstration data by model-based expert planners with a UR5 arm in PyBullet~\cite{coumans2021}. There are three statically mounted cameras~($M=3$) overlooking the tabletop as shown in Fig.~\ref{fig:overview}: one positioned 45$\degree$ downward from the front, one 50$\degree$ downward from the anti-diagonal perspective and one 50$\degree$ downward from the diagonal perspective, referred to as the front, left and right cameras respectively.
\revise{For each camera, we adopt the same camera intrinsics as those of Intel RealSense L515.} 
Our object models are from GraspNet-1Billion~\cite{fang2020graspnet}. 

{\bf Data Collection.} For both pick and place, we collect data from 5k episodes, among which the success steps are recorded as demonstrations. This results in around 6.5k successful samples in total, with approximately 3.4k for pick and 3.1k for place. During data collection of pick, 15 objects are randomly dropped into the workspace to form a cluttered scene, and the model-based pick expert planner chooses the nearest grasp of the target objects. For placement, to ensure adequate space, there are 8 objects in the workspace whose center positions are at least 0.1m from one another. The model-based place expert planner identifies the valid place region based on the reference object and the relation, and then randomly chooses a place within this region.

{\bf Visual Representations.} We employ the checkpoint of MaskCLIP ViT-L/14 to generate visual features and crop the raw image into 12 sub-images for more fine-grained features. We exclude the table points from the 3D representations for pick tasks while retaining them for place tasks. This is because the policy does not require the feature information of the table for pick action planning, and the filtering helps the policy focus on the objects. Specifically, table points are removed by height filtering of the point cloud in world coordinates. 

{\bf Training Settings.} We adopt the transformer architecture of the text encoder in \cite{radford2021learning}, with width of 768, head of 8, and layer of 1. The action decoder is a 3-layer MLP. The network parameters of MaskCLIP and action models are fixed during training. The policy is trained through cross-entropy loss for 200 epochs. During fine-tuning, the policy is trained with only 100 multi-labeled place data using binary cross-entropy loss for 200 epochs, consuming around 2 minutes. 

{\bf Hyperparameters.} We set the sample number $N=500$. For the action candidate number $L$, we sample 6 place poses for each object~(3 for ``on'' relation and 3 for ``around'' relation), while for pick, $L$ depends on the output of GraspNet. We use $\alpha=0.2$ during policy adaptation.

More implementation details can be accessed in Appendix.
\section{Experiments}
\label{sec:exp}
In this section, we carry out a series of experiments to evaluate our policy. The goals of the experiments are: 1) to validate the effectiveness of our policy in both language-conditioned pick and place tasks in clutter; 2) to demonstrate the efficiency of our policy; 3) to validate the zero-shot generalization performance of our policy on unseen objects and language instructions; 4) to test whether our policy can successfully transfer to the real world. 

\begin{figure}[t]
  \centering
  \includegraphics[width=\linewidth]{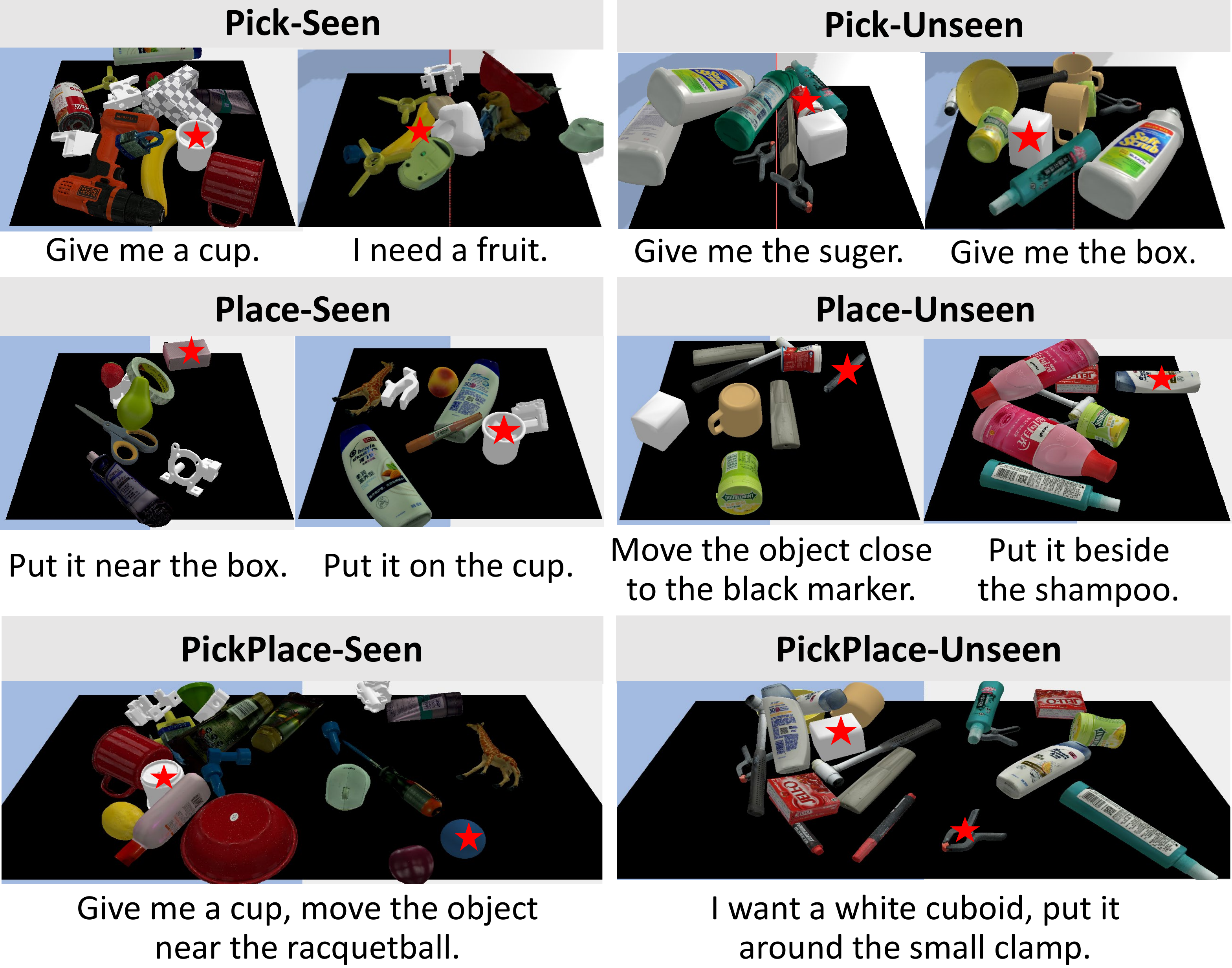}
  \vspace{-0.5cm}
  \caption{Example test cases in simulation. The target objects and reference objects are labeled with stars.}
  \label{fig:simulation_cases}
  \vspace{-0.5cm}
\end{figure}

\subsection{Experimental Setup}
{\bf Test Settings.} We first conduct test experiments in simulation with a series of test cases, which can be categorized into three folds: pick, place, and pick-n-place. Each category includes cases of arrangements with both seen and unseen objects during A$^2$ training. 
\revise{Specifically, seen objects are those that appear in the training set, while unseen objects are novel instances that are not observed during training.} 
For place, some cases of unseen objects pair with unseen relations. Example cases are visualized in Fig.~\ref{fig:simulation_cases}. For pick, each case contains 15 objects to form adversarial clutter where the robot might need to grasp away other obstacles for target grasping. All pick policies are evaluated on $x\!\!=\!\!10$ arrangements of seen objects and $x\!=\!5$ of unseen objects. For place, each case contains 8 objects to preserve some free space surrounding the reference object for placement. For place policies, performances are tested on $x\!\!=\!\!20$ arrangements of seen objects and $x\!\!=\!\!10$ of unseen objects. Given that placement is a one-step task, we add variances to action candidates in each test run of each case to evaluate robustness. For pick-n-place, we test policies with $x\!\!=\!\!8$ seen objects cases and $x\!\!=\!\!4$ unseen objects cases. Note that in these cases, we divide the workspace into a pick workspace~(left) and a place workspace~(right). To determine the success of target grasping, we use environment feedback in simulation. In real world, target is regarded as grasped if CLIP similarity of language and grasped object crop~(filtered by depth) exceeds a threshold.

{\bf Evaluation Metrics.} We evaluate methods with a series of test cases. Each contains $y\!=\!15$ runs, measured with 2 metrics:
\begin{itemize}
    \item {\bf Task Success Rate}: the average percentage of task success rate over $y$ test runs. For pick, if the robot picks up the target object within 8 action attempts, the task is considered successful and completed. For place, the robot succeeds if placing the object in the correct region with 1 action attempt. For pick-n-place, the robot should simultaneously succeed in both pick and place tasks.
    \item {\bf Planning Steps}: the average pick or pick-n-place number per task completion. Note that this metric is only evaluated in the categories of pick and pick-n-place. 
\end{itemize}

\subsection{Baselines}
We compare the performances of our policy A$^2$ to various baselines, including both modular systems and classic end-to-end policies. For modular systems, we compare to neural field based pick policies, object-centric pick and place polices, and 3D visual grounding pick and place policies.

{\bf Neural Field based Pick Policies.} These include LERF-TOGO~\cite{rashid2023lerftogo} and GraspSplats~\cite{ji2024graspsplats}. LERF-TOGO~\cite{rashid2023lerftogo} is a NeRF-based method that distills feature fields from CLIP~\cite{radford2021learning}, while GraspSplats~\cite{ji2024graspsplats} reconstructs 3D feature fields from CLIP by 3D Gaussian Splatting~\cite{kerbl20233d}. In the experiments, we train the feature fields on each step of action planning at test time. With the feature fields, they first locate the target object from language instructions, and select the corresponding grasp from GraspNet~\cite{fang2020graspnet} generated grasps. We follow the number of camera viewpoints in their papers to guarantee a fair comparison. Specifically, we add a circle of camera viewpoints around the workspace to provide sufficient information. LERF-TOGO trains its feature field with 53 posed RGB images, while GraspSplats uses 23, and both of the inputs include the 3 RGB-D images used by our method. 

{\bf Object-centric Pick and Place Policies.} There are two object-centric pick policies. VLG~\cite{xu2023joint} leverages object-centric representation to jointly model vision, language, and action information. ThinkGrasp~\cite{qian2024thinkgrasp} is an approach that develops a vision-language system with GPT4o~\cite{2024GPT4o} to plan the object grasp sequence, followed by object segmentation and grasp planning. For placement, we implement a method similar to \cite{xu2023object}, namely VLP, which grounds reference objects and spatial relations respectively. For a fair comparison, CLIP is not fine-tuned in VLP as in \cite{xu2023object}.

{\bf 3D Visual Grounding Pick and Place Policies.} We implement variant methods that directly conduct visual grounding using our 3D visual representations, named A$^2$-G-Pick and A$^2$-G-Place for pick and place tasks respectively. These methods select the action nearest to the region with the highest average similarity of K-nearest neighbors~(K=0.05$M$). In addition, A$^2$-G-Pick can combine with A$^2$-G-Place as a 3D visual grounding pick-n-place policy, denoted as A$^2$-G. 

{\bf 3D End-to-end Pick and Place Policies.} We compare to 3D end-to-end policies Act3D~\cite{gervet2023act3d}, RVT-2~\cite{goyal2024rvt2} and 3D Diffuser Actor~\cite{ke20243d}, which leverage multi-view CLIP features to predict 3D actions. We use pre-trained models of Act3D and RVT-2, as well as the models trained on our data~(referred to as Act3D$^\dag$, RVT-2$^\dag$, and 3D Diffuser Actor$^\dag$) for evaluation. Note that the setting of the pre-trained model of 3D Diffuser Actor is distinct from our setting, thus cannot be directly employed.

\subsection{Comparison to Baselines}

{\bf Pick.} Results in Table~\ref{table:baseline} indicate that our policy outperforms all baselines. Although LERF-TOGO and GraspSplats can obtain fine-grained scene representations via time-consuming~($\textgreater$1min) test-time training, the grounding accuracy is hindered in clutter, leading to cascaded errors in action planning. Therefore, they demonstrate unsatisfactory performances. Other methods support real-time inference. VLG gets object awareness by incorporating object-centric representation, but suffers from detection noise, resulting in lower task success rates. ThinkGrasp utilizes GPT-4o as the planner based on the object-centric crops, which inherits the reasoning capability of LLM. Nevertheless, it operates in a stage-by-stage manner, affected by the accuracy of segmentation and LLM planning, calling for more planning steps for some fuzzy concepts. A$^2$-G-Pick relies on the similarity cloud for grounding, and ignores the probability of moving away other obstacles. In contrast, action prior alignment enables our policy to directly score actions based on task-relevant vision-language features. In this way, our policy avoids over-reliance on accurate visual representations and can remove obstacles for target grasping.

{\bf Place.} We show the performances of place with seen and unseen objects in Table~\ref{table:baseline}, further demonstrating the advantages of our action prior alignment paradigm. The performance of VLP depends heavily on the capability of CLIP, which frequently fails when facing similar visual information or text words. A$^2$-G-Place struggles to distinguish ``in'' and ``around'' relation, as it directly grounds the highest point that fits both requirements of reference and relation. 

{\bf Pick-n-Place.} As shown in Table~\ref{table:baseline}, Act3D, RVT-2 fail in all cases when employing their pre-trained models, revealing poor generalization to novel objects, backgrounds, and camera viewpoints. Even when trained on our dataset, Act3D$^\dag$, RVT-2$^\dag$, and 3D Diffuser Actor$^\dag$ still struggle to acquire the necessary information to complete tasks, likely due to insufficient data quantity. By further leveraging action foundation priors and aligning them based on zero-shot vision-language priors, our policy achieves higher efficiency and generalization.

{\bf Generalization.} All policies are tested with objects seen and unseen during A$^2$ training. Overall, our policy achieves the highest task success rates in unseen objects, particularly excelling in pick tasks. Thanks to our design of action prior alignment desgin, we effectively preserve the generalization capabilities of the foundation models to a large extent.

\begin{table}[t]
\caption{Simulation Results on All Categories and Arrangements}
\label{table:baseline}
\centering
\begin{tabular}{cccc}
\toprule
Category & Method & Seen & Unseen \\
\midrule
\multirow{6}{*}{Pick} & LERF-TOGO & 83.3/3.37& 76.0/\underline{2.01}\\
& GraspSplats & 58.0/\bf{2.05} & 37.3/\bf{1.67}\\
& VLG & 74.3/4.11 & 78.7/3.98 \\
& ThinkGrasp & \underline{84.7}/\underline{2.55} & 57.3/4.11 \\
& A$^2$-G-Pick & 83.3/3.78 & \underline{84.7}/{3.85} \\
& A$^2$ & {\bf 95.3}/\underline{2.55} & {\bf 97.3}/2.57 \\
\midrule
\multirow{4}{*}{Place} & VLP & 40.0 & 20.0 \\
& A$^2$-G-Place & 32.3 & 29.3 \\
& A$^2$ & \bf{89.3} & \underline{74.0} \\
& A$^2$-PA & \underline{89.0} & \bf{76.0} \\
\midrule
\multirow{8}{*}{Pick-n-Place} & Act3D & 0.0/-- & 0.0/-- \\
& RVT-2 & 0.0/-- & 0.0/-- \\
& Act3D$^\dag$ & 0.0/-- & 0.0/-- \\
& RVT-2$^\dag$ & 0.83/4.00 & 0.0/-- \\
& 3D Diffuser Actor$^\dag$ & 1.67/6.13 & 0.0/-- \\
& A$^2$-G & 30.7/\underline{2.42} & 28.3/\bf{2.00} \\
& A$^2$ & \underline{87.5}/2.45 & \underline{71.7}/\underline{3.02} \\
& A$^2$-PA & \bf{91.3}/\bf{2.06} & {\bf 76.7}/3.22 \\
\bottomrule
\end{tabular}
\begin{tablenotes}
\footnotesize
\item * Metrics of pick and pick-n-place are presented as Task Success Rate / Planning Steps.
\end{tablenotes}
\vspace{-0.5cm}
\end{table}

\begin{table}[t]
\caption{Inference Time of Different Policies}
\label{table:inference-time}
\centering
\begin{tabular}{ccccc}
\toprule
Method & Inference Time\\
\midrule
LERG-TOGO & $\sim$5.2min\\
GraspSplats & $\sim$80.0s\\
ThinkGrasp & $\sim$7.5s\\
RVT-2 & $\sim$2.0s\\
Act3D & $\sim$1.5s\\
3D Diffuser Actor & $\sim$3.0s\\
A$^2$ & $\sim$1.0s \\
\bottomrule
\end{tabular}
\vspace{-0.2cm}
\end{table}

{\bf Inference Time.} We report the inference times of policies in Table~\ref{table:inference-time}. All the policies are run on an RTX 4090 and AMD EPYC 9354 CPU. Obviously, neural field based policies~(LERG-TOGO, GraspSplats) are time-consuming due to test-time training of feature fields. By incorporating GPT-4o, ThinkGrasp avoids test-time training, but is still limited by the complex stage-by-stage process. Other policies show faster inference speeds, including Act3D, RVT-2, 3D Diffuser Actor, and A$^2$. Among them, our policy can predict an action key pose in approximately 1.0s. Note that all policies predict key poses, followed by the same trajectory planning algorithm.

\subsection{Ablation Studies}
\label{sec:ablation}

\begin{figure}[t]
  \centering
  \includegraphics[width=\linewidth]{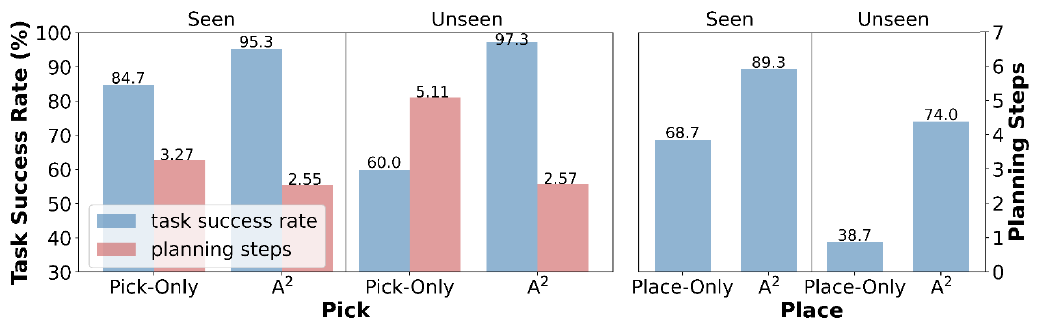}
  \vspace{-0.6cm}
  \caption{Ablation studies of shared policy.}
  \label{fig:ablation}
  \vspace{-0.4cm}
\end{figure}

We conduct extensive ablation studies to elucidate the effectiveness of individual designs within our method. For a fair comparison, all the learning-based methods are trained with the same process.

{\bf Shared Policy.} We first test the effectiveness of the shared policy for pick and place. Let Pick-Only denote the policy trained only with pick samples and Place-Only as the policy trained only with place samples. Results in Fig.~\ref{fig:ablation} demonstrate the shared policy boosts performances in both tasks by a large margin. This indicates strong commonalities between pick and place tasks, as both require focus on or around the target region. Training a shared policy for both tasks enables mutual reinforcement between the two skills: for picking in clutter, exposure to place data encourages the robot to move obstacles around the target, while for placing, pick data enhances focus on the reference object.

{\bf Policy Adaptation.} To validate the policy adaptation scheme, we compare the performances of our policy before~(A$^2$) and after adaptation~(namely A$^2$-PA) with only 100 multi-labeled place samples. It can be seen from Table~\ref{table:baseline} that our policy adaptation scheme can improve the generalization performances in both place and pick-n-place tasks. It is interesting to note that A$^2$-PA outcomes fewer planning steps for pick-n-place tasks involving seen objects, suggesting that policy adaptation on place data also facilitates the efficiency of picking. This might be because fine-tuning with multi-labeled data brings multi-modal characteristics, better fitting the true action distribution. And multi-modality is a commonality of pick and place actions.

\begin{table}[t]
\caption{Ablation Studies of Different Policy Adaptations}
\label{table:abaltion_pa}
\centering
\begin{tabular}{cccccc}
\toprule 
{RL} & {\revise{IL}} & {Res} & {Data} & {Seen} & {Unseen} \\
\midrule 
$\checkmark$ & & & 1500 & 36.7 & 28.0 \\
$\checkmark$ & & $\checkmark$ & 1500 & 89.7 & 68.0\\
& $\checkmark$ & & 100 & 56.3 & 19.7\\
& $\checkmark$ & $\checkmark$ & 100 & 89.0 & 76.0\\
\bottomrule
\end{tabular}
\vspace{-0.2cm}
\end{table} 

\begin{table}[t]
\caption{Ablation Studies of Network Architecture}
\label{table:abaltion}
\centering
\begin{tabular}{p{0.2cm}p{0.2cm}cccccc}
\toprule
\multirow{2}{*}{TE} & \multirow{2}{*}{LE} & \multirow{2}{*}{RoPE} & \multirow{2}{*}{RGB} & \multicolumn{2}{c}{Pick} & \multicolumn{2}{c}{Place} \\
\cline{5-8}
& & & & Seen & Unseen & Seen & Unseen \\
\midrule
% & & $\checkmark$ & 84.7/3.27 & 60.0/5.11 & & \\
% & & $\checkmark$ & & & 68.7 & 38.7 \\
& & & $\checkmark$ & 90.7/2.41 & 80.0/3.20 & 69.0 & 52.0 \\
& & $\checkmark$ & & 60.0/4.58 & 73.3/3.55 & 26.7 & 42.7 \\
& & $\checkmark$ & $\checkmark$ & 95.3/2.55 & 97.3/2.57 & 89.3 & 74.0 \\
\midrule
$\checkmark$ & & $\checkmark$ & $\checkmark$ & 92.7/2.84 & 78.7/3.06 & 74.0 & 39.3 \\
& $\checkmark$ & $\checkmark$ & $\checkmark$ & 94.0/2.48 & 88.0/2.39 & 71.3 & 43.3 \\
\bottomrule
\end{tabular}
\begin{tablenotes}
\footnotesize
\item * Metrics of pick are presented as Task Success Rate / Planning Steps.
\end{tablenotes}
\vspace{-0.4cm}
\end{table}

\begin{figure}[t]
  \centering
  \includegraphics[width=\linewidth]{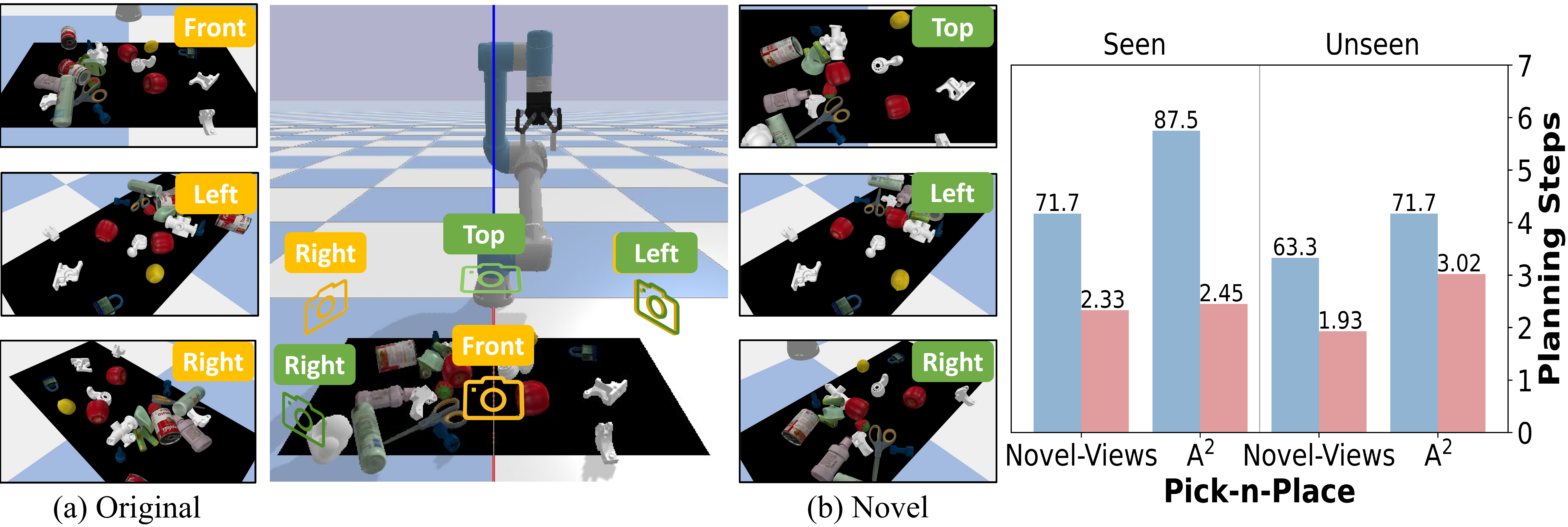}
  \vspace{-0.5cm}
  \caption{Ablation studies of (a) original camera viewpoints and (b) novel camera viewpoints.}
  \label{fig:diff_views}
  \vspace{-0.3cm}
\end{figure}

\begin{table}[t]
\centering
% \color{blue}
\caption{Scaling to More Objects and Data}
\label{table:scale}
\begin{tabular}{ccccc}
\toprule
 & \multicolumn{2}{c}{Pick} & \multicolumn{2}{c}{Place} \\
\cline{2-5}
 & Seen & Unseen & Seen & Unseen \\
\midrule
% A$^2$ & 95.3/2.55 & 97.3/2.57 & 89.3 & 74.0 \\
% \hline
A$^2$-2\#O & 81.3/3.89 & 78.7/3.18 & 81.3 & 56.7 \\
A$^2$-2\#D & 97.3/2.36 & 97.3/2.31 & 92.0 & 79.3 \\
% \hline
% A$^2$-P & 60.0/4.58 & 73.3/3.55 & 26.7 & 42.7 \\
\bottomrule
\end{tabular}
\begin{tablenotes}
\footnotesize
\item * Metrics of pick are presented as Task Success Rate / Planning Steps.
\end{tablenotes}
\vspace{-0.4cm}
\end{table}

\begin{figure*}[t]
  \centering
  \includegraphics[width=\linewidth]{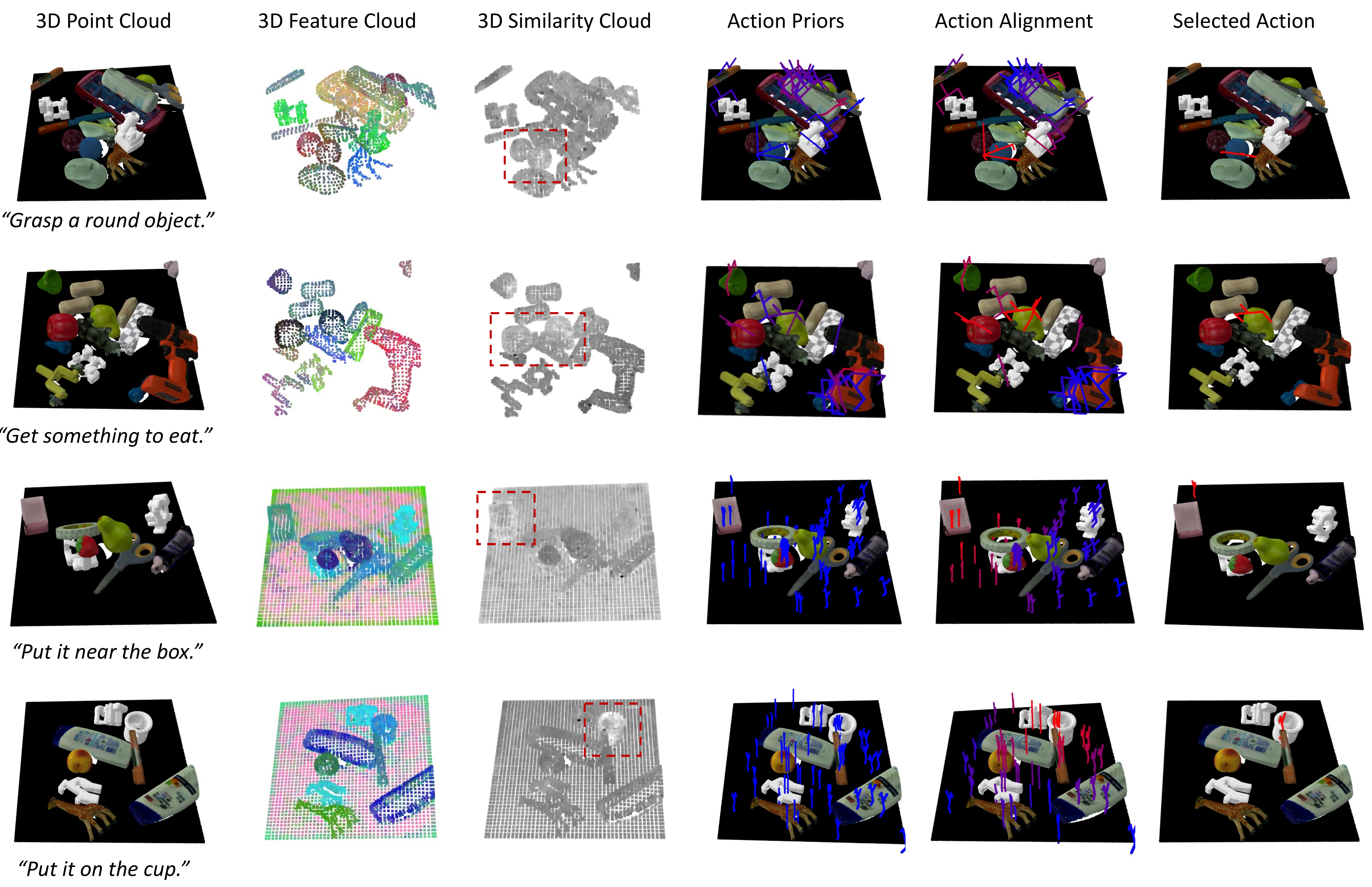}
  \vspace{-0.7cm}
  \caption{Case studies. For each case, we show the 3D representations~({\it i.e.} 3D point cloud, 3D feature cloud, and 3D similarity cloud), the action priors from action foundation models, the alignment results, and the final selected action. Notably, in the similarity cloud, regions with high similarity are highlighted with red rectangles. For each action, the labeled color indicates the action probability, with the color shifting toward red as the probability increases.}
  \label{fig:case_study}
  \vspace{-0.2cm}
\end{figure*}

\begin{figure*}[t]
  \centering  \includegraphics[width=\linewidth]{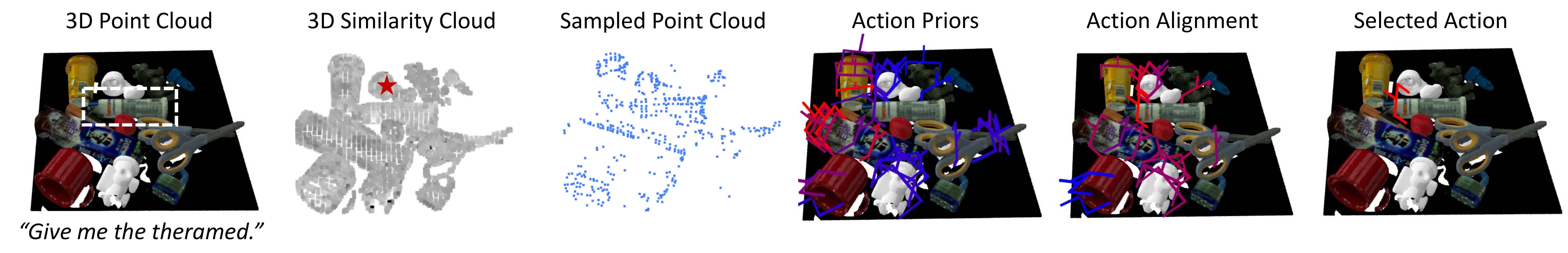}
  \vspace{-0.7cm}
  \caption{Case visualization where the visual grounding fails, yet our policy selects the correct grasp via alignment. The white rectangle in the 3D point cloud marks the target, while the red star in the similarity cloud marks the direct visual grounding result. For each action, the labeled color indicates the action probability, with the color shifting toward red as the probability increases.}
  \label{fig:case_study_}
  \vspace{-0.4cm}
\end{figure*}

\begin{figure}[t]
  \centering
  % \color{blue}
  \includegraphics[width=\linewidth]{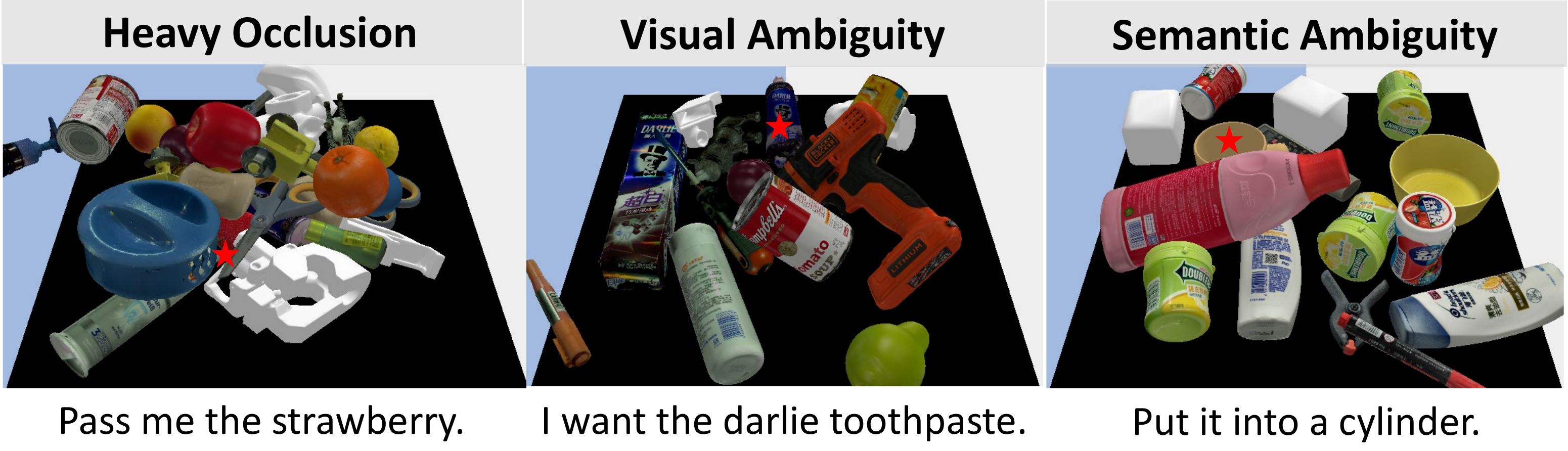}
  \vspace{-0.5cm}
  \caption{Example failure modes, including heavy occlusion, visual ambiguity, and semantic ambiguity.}
  \label{fig:case_fail}
  \vspace{-0.6cm}
\end{figure}

\begin{figure*}[t]
  \centering
  \includegraphics[width=\linewidth]{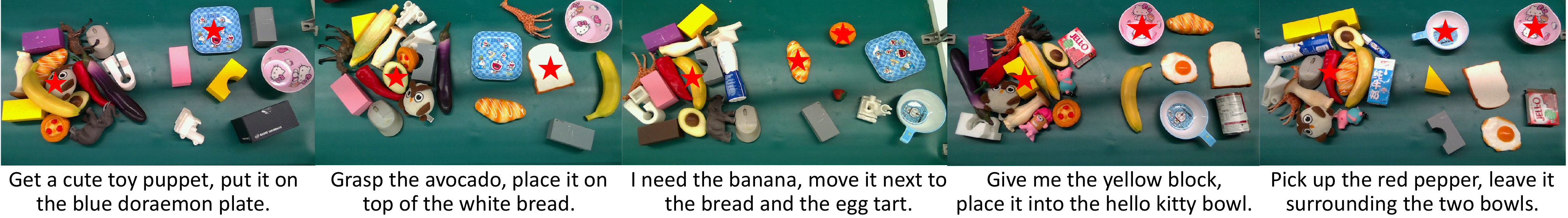}
  \vspace{-0.6cm}
  \caption{Test cases in real world. Each case contains 21$\sim$22 objects that are mostly unseen during training. Target or reference objects are labeled with stars.}
  \label{fig:full_real_cases}
  \vspace{-0.4cm}
\end{figure*}

{\bf Different Policy Adaptations.} 
\revise{We compare different policy adaptation strategies, including learning paradigms~(reinforcement learning~(RL), imitation learning~(IL)), data amounts, and network architectures~(with or without residual blocks). For RL, we adopt Soft Actor-Critic~\cite{haarnoja2018soft,christodoulou2019soft}. In the ``Res'' variant, only the residual block is updated. Otherwise, the entire network is fine-tuned. As shown in Table~\ref{table:abaltion_pa}, residual-only fine-tuning significantly outperforms full fine-tuning, as it preserves pretrained knowledge while enabling efficient adaptation. In contrast, full fine-tuning may harm generalization, likely due to catastrophic forgetting. With residual blocks, IL and RL perform similarly, but RL requires more data and generalizes worse. This is likely because the one-step nature of place tasks limits RL’s advantage in sequential learning.

}

{\bf Network Architecture.} We compare our method with some variant methods to evaluate the architecture design. Testing results are shown in Table~\ref{table:abaltion}. 
Removing RoPE causes notable drops: $17.3\%$ for unseen pick tasks and over $20\%$ for place tasks, highlighting its importance for generalization. 
In addition, using sampled point features as values in cross-attention instead of image features degrades performance, showing the benefit of foundation model features for effectiveness and generalization.
Also, adding task-specific embeddings~(TE) to action features harms performance, likely by hindering shared representations between pick and place. 
Finally, directly feeding the language embedding~(LE) into cross-attention instead of weighting visual features with similarities weakens CLIP priors and reduces success rates.

{\bf Novel Camera Viewpoints.} We vary the camera viewpoints at test time and present the results in Fig.~\ref{fig:diff_views}. It is shown that our policy can generalize to novel camera viewpoints. This benefits from our zero-shot 3D representation, which does not impose strict constraints on camera viewpoints. As a result, our policy remains effective and practical for deployment in new scenarios with varying camera configurations, offering greater flexibility for real-world applications.

\revise{
{\bf Generalization to More Objects.} To demonstrate the generalization to more object distractors and denser clutter, we double the number of objects for testing as A$^2$-2\#O in Table~\ref{table:scale}. Notably, there is no retraining of policy. Results show that our policy outperforms most baselines~(tested with original object number) even with double objects, validating effectiveness in more complex settings.

{\bf Scaling to More Data.} We double the training data to test the scalability, and report results as A$^2$-2\#D in Table.~\ref{table:scale}, which verifies effective improvements in both pick and place tasks when scaling to more data.
}

{\bf Case Studies.} Fig.~\ref{fig:case_study} shows several cases to illustrate the 3D representations, action priors, and alignment results of our policy. Given language instructions, the similarity cloud can highlight the task-relevant regions, and our policy aligns action priors based on these representations. Fig.~\ref{fig:case_study_} further shows a case where visual grounding alone fails, but our alignment still enables correct grasp selection through alignment. This indicates that while we take similarity-sampled points, we evaluate actions guided by visual grounding rather than being determined by it. 
\revise{Fig.~\ref{fig:case_fail} visualizes some typical failure modes, including heavy occlusion and visual ambiguity of target objects, as well as semantic ambiguity in language instruction, {\it i.e} ambiguous word ``cylinder''.
}

\subsection{Real-world Experiments}
\begin{figure}[t]
  \centering
  % \color{blue}
  \includegraphics[width=0.9\linewidth]{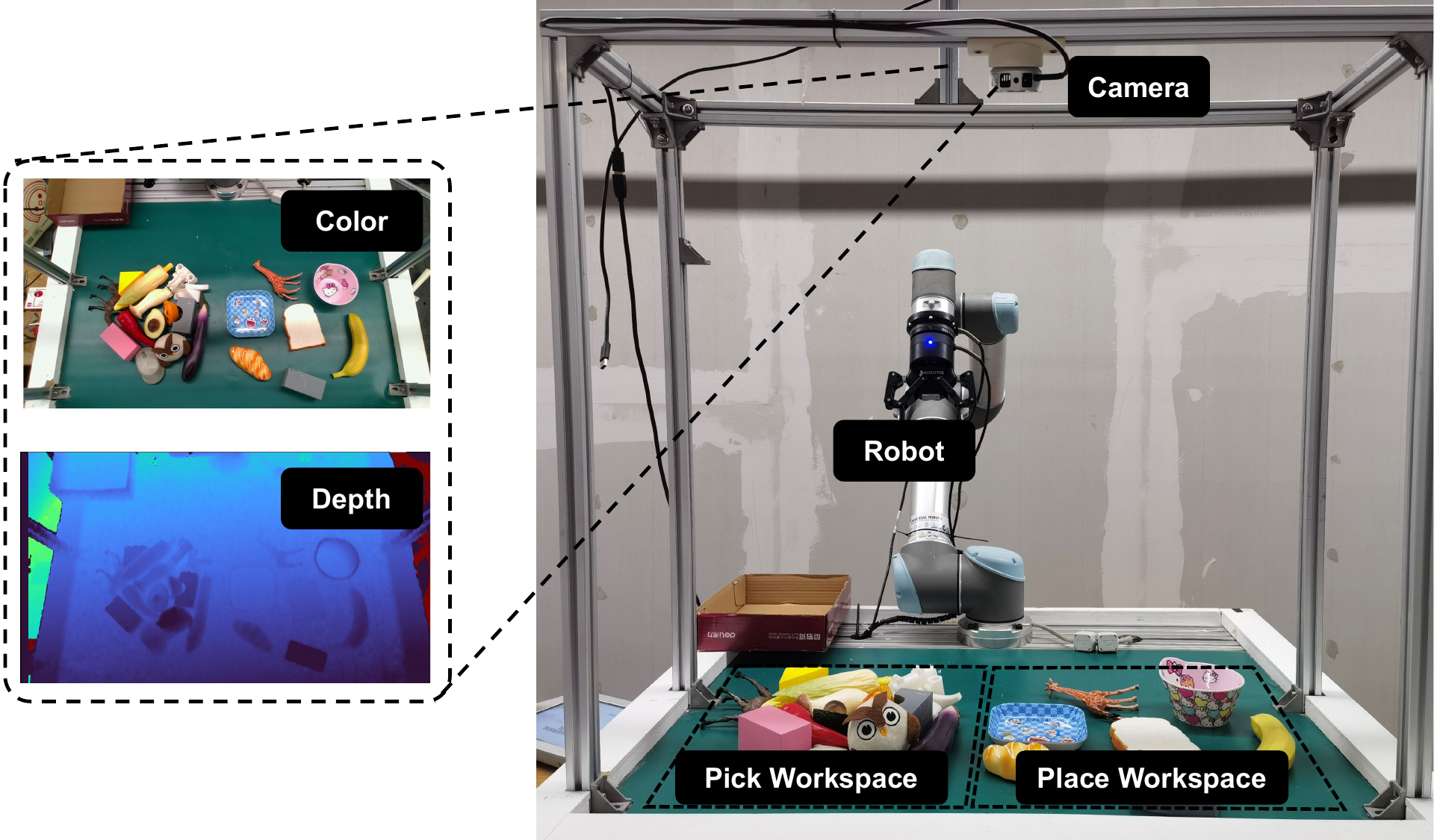}
  \vspace{-0.1cm}
  \caption{Real-world platform, which involves a UR5 robot arm equipped with a ROBOTIQ-85 gripper, and an Intel RealSense L515 camera.}
  \label{fig:real_world}
  \vspace{-0.2cm}
\end{figure}
\begin{figure*}[t]
  \centering  \includegraphics[width=\linewidth]{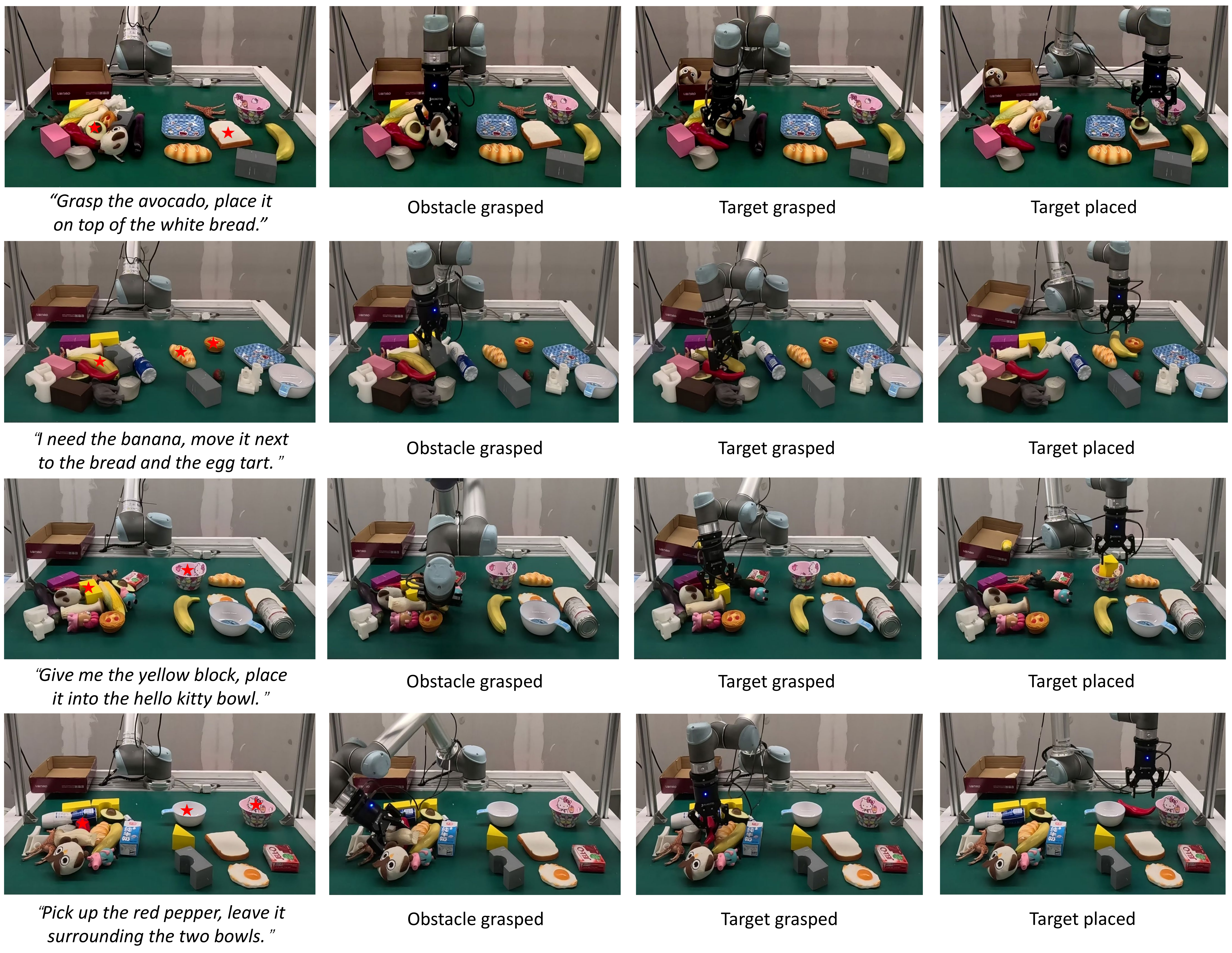}
  \vspace{-0.6cm}
  \caption{Example testing sequences. The camera viewpoint and most of the objects are unseen during training. Taking the language instructions for pick and place, our policy is able to gradually remove obstacles, grasp the target object, and finally place it at the target location.}
  \label{fig:real_sequence}
  \vspace{-0.4cm}
\end{figure*}

{\bf Experiment Setup.} In this section, we evaluate our policy in real-world settings. Fig.~\ref{fig:real_world} shows our real-world setup, which involves a UR5 robot arm equipped with a ROBOTIQ-85 gripper, and an Intel RealSense L515 capturing RGB-D images at a resolution of 1280$\times$720. Notably, the camera viewpoint in the real-world setup is unseen during training. \revise{We use a single camera to evaluate generalization under limited-view settings while avoiding depth interference caused by multiple sensors.} 
The workspace is divided into pick and place workspaces, where the robot is supposed to grasp the target object within the pick workspace, and place it within place workspace. Our test cases include 5 scenarios shown in Fig.~\ref{fig:full_real_cases}. Each of them contains 21$\sim$22 objects that are mostly unseen during training. There are in total of 38 objects for real-world testing, including 10 seen objects and 28 unseen objects. 

At inference, the policy first receives the pick language instruction, and plans actions upon the point cloud within the grasp workspace. Once the target object is grasped, the place language instruction is fed into the policy for action planning, with the point cloud within the place workspace. For the place action model, we employ a pre-trained model to generate object region proposals~\cite{zhou2023open}, which is trained on data from GraspNet-1Billion~\cite{fang2020graspnet} with $mAP=70.70$ for seen objects and $mAP=34.53$ for unseen objects.

{\bf Comparison to Baselines.} We compare our policy with A$^2$-G, as it performs better in simulated experiments. We test 10 runs for each case, in a total of 50 times testing. All the policies are transferred from simulation to the real world without additional training. Test results are reported in Table~\ref{table:real}. In general, our policy achieves much better performance in task success rate. Though A$^2$-G demonstrates fewer planning steps, it gets a low task success rate at 56$\%$. This is due to the fact that A$^2$-G cannot afford errors in visual grounding. Instead, by building an action architecture, our policy assesses the probabilities of feasible actions conditioned on vision-language cues, reducing reliance on visual grounding accuracy. By further injecting multi-modality characteristics, A$^2$-PA improves performances in both task success rate and planning efficiency.

{\bf Generalization.} The results in Table~\ref{table:real} further verify the generalization capability of our policy to camera number, camera viewpoints, and novel objects. This is not merely due to our application of foundation models, but also because our alignment design effectively integrates the priors of multiple foundation models through a lightweight network, all while preserving the knowledge embedded in the pre-trained models.

{\bf Example Sequences.} Fig.~\ref{fig:real_sequence} illustrates some execution sequences in real-world experiments. In a cluttered environment, by scoring the candidate actions through alignment, our policy displays the ability to gradually remove obstacle objects, grasp the target object, and finally place it at the specified location.

\begin{table}[t]
\caption{Real-world Results on Test Arrangements}
\label{table:real}
\centering
\resizebox{\linewidth}{!}{
\begin{tabular}{ccccccc}
\toprule
Method & Case 1 & Case 2 & Case 3 & Case 4 & Case 5 & Average \\
\midrule
A$^2$-G  & 40.0/4.75 & 80.0/2.00 & 60.0/3.00 & 80.0/2.88 & 20.0/4.50 & 56.0/\textbf{3.04} \\
A$^2$    & 70.0/4.29 & 80.0/5.50 & 80.0/2.63 & 80.0/2.63 & 70.0/4.86 & \underline{76.0}/3.95 \\
A$^2$-PA & 70.0/3.43 & 80.0/4.13 & 80.0/3.13 & 90.0/2.78 & 80.0/5.00 & \textbf{80.0}/\underline{3.68} \\
\bottomrule
\end{tabular}}
\vspace{-0.4cm}
\end{table}
\section{Conclusion}
\label{sec:conclusion}

We present A$^2$, an action prior alignment method for language-conditioned pick and place in clutter. We propose to align action priors based on 3D vision-language priors. Using foundation models, we construct a 3D zero-shot visual representation, and generate candidate actions that provide feasible action patterns. Conditioned on these foundation priors, we conduct alignment by learning one attention layer to score the candidate actions for downstream tasks. Experiments show that our policy requires less training data, supports fast adaptation, and achieves higher task success rates with fewer planning steps than other methods. Additionally, our method demonstrates zero-shot generalization to unseen objects and language instructions, effectively transferring from simulation to the real world. 

\textbf{Limitations and future work.} In this paper, we evaluate A$^2$ in language-conditioned pick and place tasks. In the current version, our policy processes pick and place instructions separately. Nevertheless, we can easily decompose a compound instruction into pick and place components by Large Language Models~(LLM) like GPT-4~\cite{2024GPT4o}, which can also determine whether the target is grasped. Besides, our policy struggles with strict place constraints in clutter. By incorporating more action foundation models~({\it e.g.} Vision-Language-Action models~\cite{o2023open,kim2024openvla,team2024octo,liu2024rdt}), we believe that our action prior alignment approach can be extended to a wider range of tasks, offering a promising direction for future research.

\bibliographystyle{IEEEtran}
\bibliography{IEEEabrv,references}
% \printbibliography

\appendix
\subsection{3D Representation Details}
Given image(s) $\mathcal{I}\!=\!\{I_i\}_{i=0,1,...,M}$ of one or more RGB-D camera(s), we extract 2D patch-level features $\mathcal{W}_i$ by MaskCLIP~\cite{zhou2022extract}, including visual patch-level features $\mathcal{W}^f_i$ and vision-language similarity information $\mathcal{W}^s_i$ denoting cosine similarities between language embeddings and $\mathcal{W}^f_i$.

We generate a 3D point cloud $\mathbf{p}$ within the workspace using the camera parameters. For each point $p_j$ of $\mathbf{p}$, we project it back to the $i$th camera viewpoint as the pixel $u_j^i$, and get its visual feature $f_j^i$ by interpolation:
\setcounter{equation}{0}
\begin{equation}
\label{eq:single_feat}
f_j^i = \mathcal{W}^f_i[u_j^i]
\end{equation}

Following \cite{wang2024d}, we compute weights for each camera according to the visibility and distance of $p_j$ relative to the $i$th camera. We denote the distance from $p_j$ to the $i$th camera viewpoint as $l_i$, and compute the depth by interpolating the corresponding depth image $I_i^d$ as $l_i^\prime=I_i^d[u_j^i]$. Then the truncated depth difference is defined as:
\begin{equation}
\label{eq:trunc_depth}
\begin{aligned}
    d_i = l_i - l_i^\prime,
    &\quad
    d_{i}' = \textrm{max}(\textrm{min}(d_i, \mu), -\mu),\\
\end{aligned}
\end{equation}
where $\mu=0.02$ represents the truncation threshold for the Truncated Signed Distance Function~(TSDF). The visibility of $p_j$ in the $i$th camera viewpoint can be represented as $v_i=\mathds{1}_{d_i<\mu}$. Here $\mathds{1}$ is the indicator function. We compute the weight for the $i$th camera viewpoint as: 
\begin{equation}
\label{eq:weights}
\beta_i = \exp{\left(\frac{\min\left( \mu - |d_i|, 0\right)}{\mu}\right)}.
\end{equation}
where $\beta_i$ decays as $|d_i|$ increases. Then, we can obtain the semantic feature $f_j$ by fusing features from $M$ camera viewpoints:
\begin{equation}
f_j = \frac{\sum_{i=1}^{M} \beta_i v_i f_j^i}{\epsilon + \sum_{i=1}^{M} v_i}
\label{eq:fuse}
\end{equation}
where $\epsilon=1\times10^{-6}$ is to avoid numeric issues. 

Similarly, we can get the similarity value $s_j$ for $p_j$ in the same way upon $\mathcal{W}^s_i$. Finally, we get a 3D feature cloud $\mathbf{f}=\{f_j\}$ indicating the visual features and a 3D similarity cloud $\mathbf{s}=\{s_j\}$ indicating the task-relevant information.

\subsection{Data Collection Details}

{\bf Language Instructions.} During each rollout of data collection, we randomly sample a language template along with keywords~(target for pick tasks, reference and relation for place tasks) to form a complete language instruction. For pick, there are five language templates: ``Give me the \{target\}'', ``I need a \{target\}'', ``Grasp a \{target\} object'', ``I want a \{target\} object'', ``Get something to \{target\}'', while there are three for place: ``Put it \{relation\} the \{reference\}'', ``Place this \{direction\} the \{reference\}'', ``Move the object \{direction\} the \{reference\}''. There is a total of 66 object models for data collection, with 36 language keywords categorized into four types: 
\revise{
concrete labels, general categories, attributes of color or shape, and functional descriptions. The four types of object label follow a 4:2:2:2
distribution, as shown in Fig.~\ref{fig:object_label}.
}
For spatial relations, there are 6 choices for ``on'' or ``around'' relations. 

\begin{figure}[t]
  \centering
  % \color{blue}
  \includegraphics[width=0.8\linewidth]{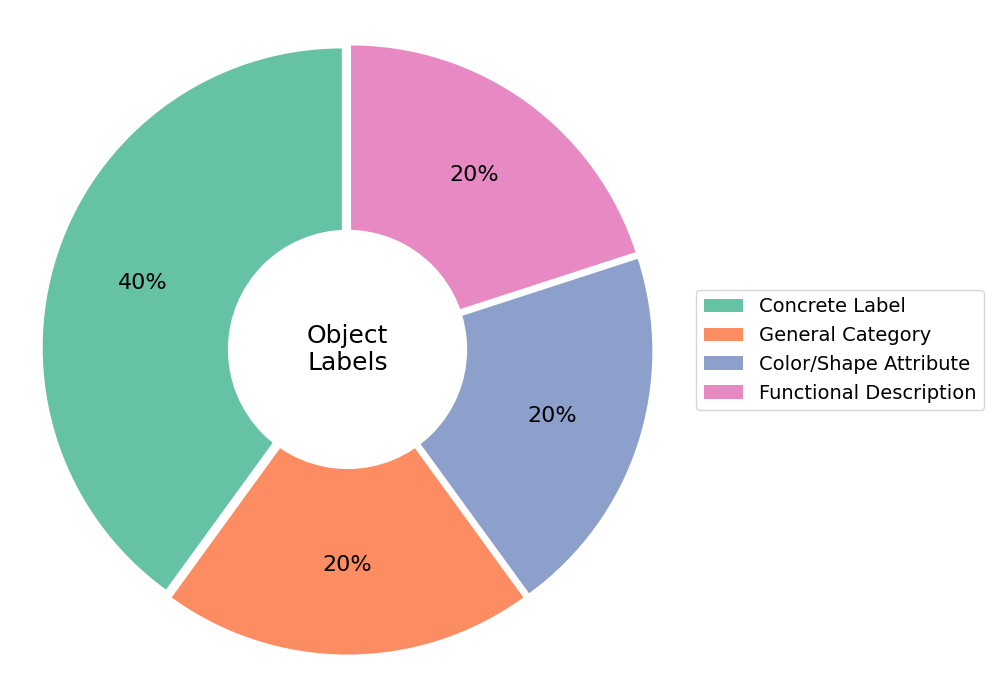}
  \vspace{-0.2cm}
  \caption{Diversity of object labels in language instructions, including four types: concrete labels, general categories, attributes of color or shape, and functional descriptions, with a distribution ratio of 4:2:2:2.}
  \label{fig:object_label}
  \vspace{-0.1cm}
\end{figure}

\begin{figure}[t]
  \centering
  \includegraphics[width=\linewidth]{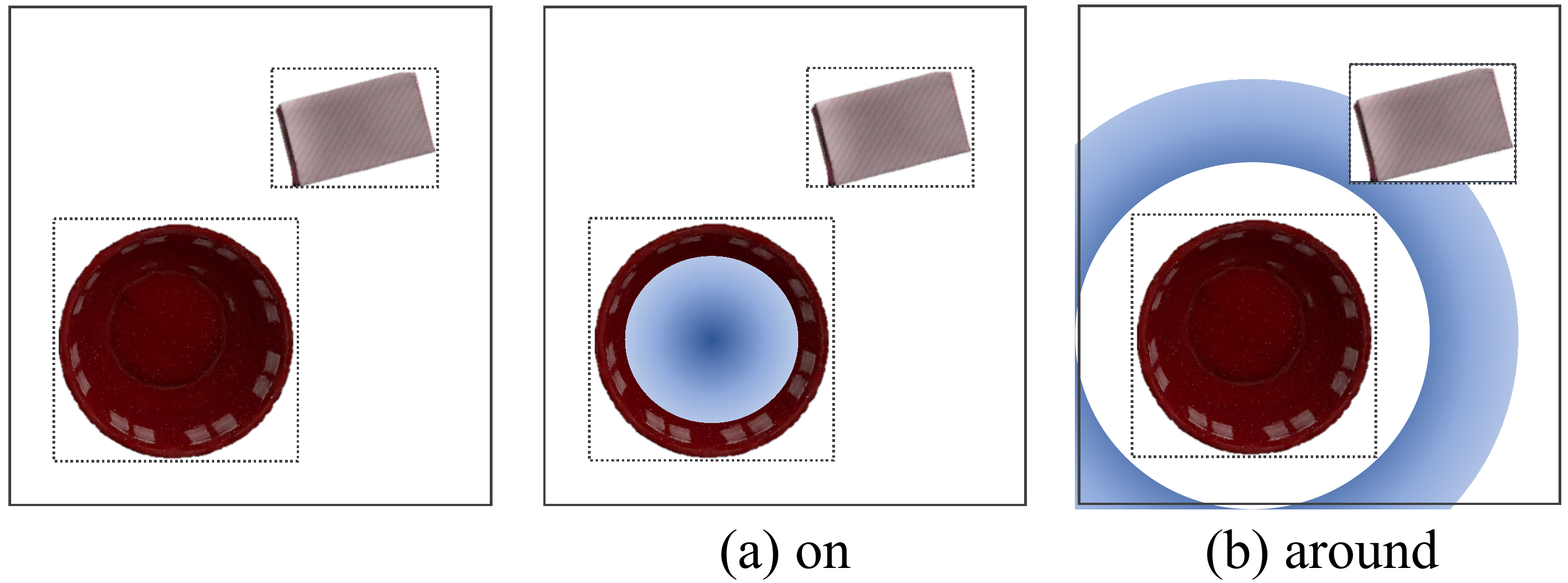}
  % \vspace{-0.2cm}
  \caption{Example generated place regions for (a) ``on'' relation and (b) ``around'' relation relative to the red bowl.}
  \label{fig:place_dist}
  \vspace{-0.1cm}
\end{figure}

{\bf Model-based Experts.} We collect data with model-based expert planners. The model-based pick expert planner selects the grasp nearest to the target objects from candidates generated by GraspNet~\cite{fang2020graspnet}. The model-based place expert planner determines valid place regions based on the reference object and the relation. Specifically, we first obtain object region proposals from the mask image in Pybullet~\cite{coumans2021}, where each pixel donates the index of the object visible in the camera. Object regions are identified as bounding boxes of pixels with the same index, and regions whose size is smaller than $5\times5$ are discarded. Then the valid place region is generated within the reference object for the ``on'' relation, or around the reference object for the ``around'' relation. Note that the generated ``around'' region should not overlap with any object regions. Fig.~\ref{fig:place_dist} shows example place regions for the ``on'' and ``around'' relations. 

{\bf Visual Representation Filtering.} We exclude the table points from the visual representations~({\it i.e.} 3D feature cloud and 3D similarity cloud) for pick tasks while retaining them for place tasks. Specifically, table points are removed by height filtering of the point cloud in world coordinates. This is because the policy does not require the feature information of the table for pick action planning, and the filtering helps the policy focus on the objects.

\revise{
{\bf Imitation Learning Setting.} Regarding Eqn.~\ref{eq:imitation}, our goal is to maximize the likelihood of the successful action $a_d$ among the candidate actions $\mathcal{A}_L(\mathcal{I}_d)$ for each demonstration $\mathcal{D} = \{\mathcal{I}_d, \mathcal{L}_d, a_d\}$. To be specific, we formulate this as a maximum likelihood estimation~(MLE) problem, which is optimized via the cross-entropy loss:
\begin{equation}
\mathcal{L}_{\text{CE}} = - \log \omega(a_d | \mathcal{I}_d, \mathcal{L}_d)
\end{equation}
\vspace{-0.2cm}
}

\begin{figure*}[t]
  \centering
  \includegraphics[width=\linewidth]{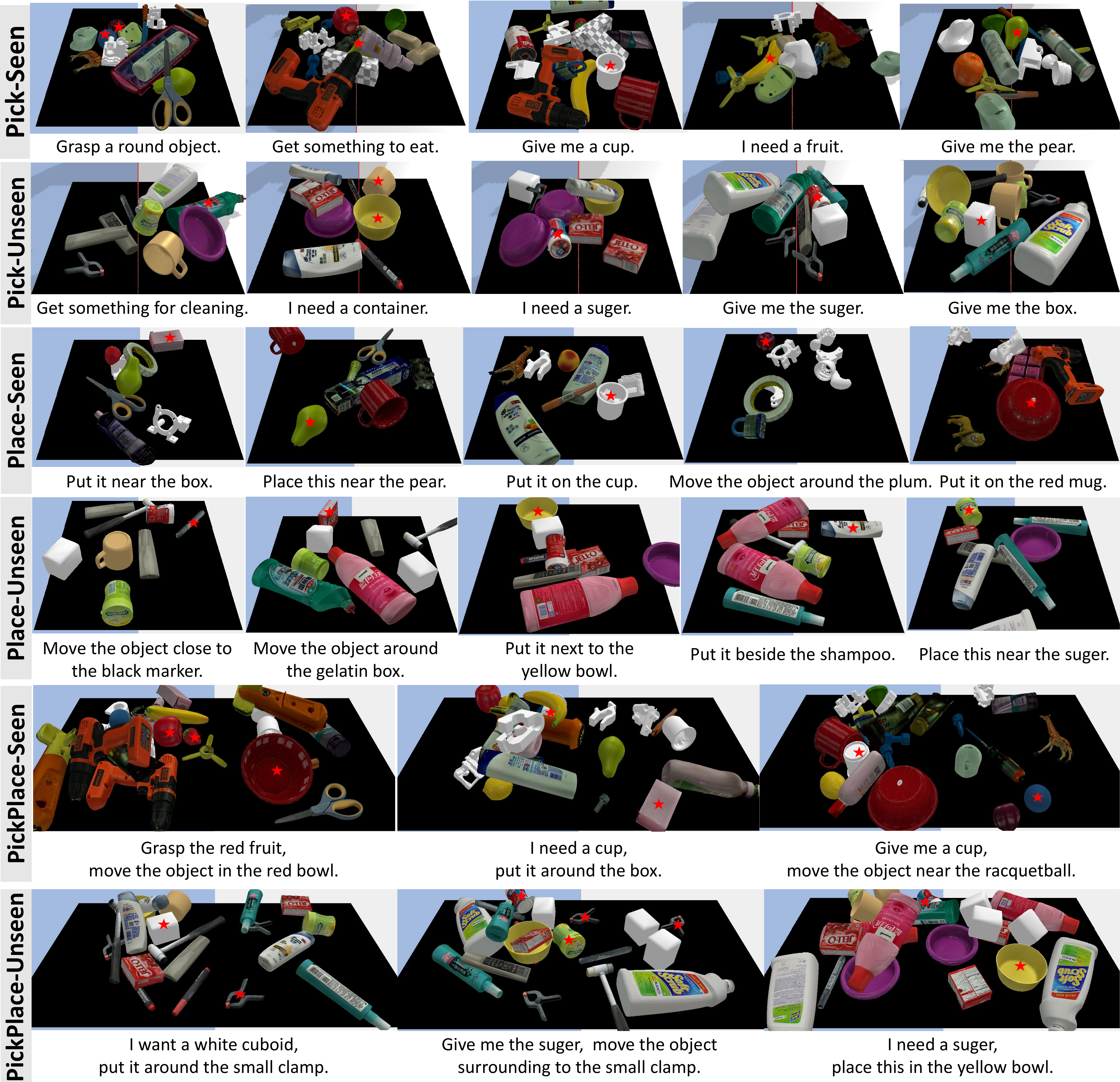}
  \caption{More example test cases in simulation. The target objects or reference objects are labeled with stars.}
  \label{fig:more_simulation_cases}
  % \vspace{-0.2cm}
\end{figure*}

\begin{figure}[t]
  \centering
  % \color{blue}
  \includegraphics[width=\linewidth]{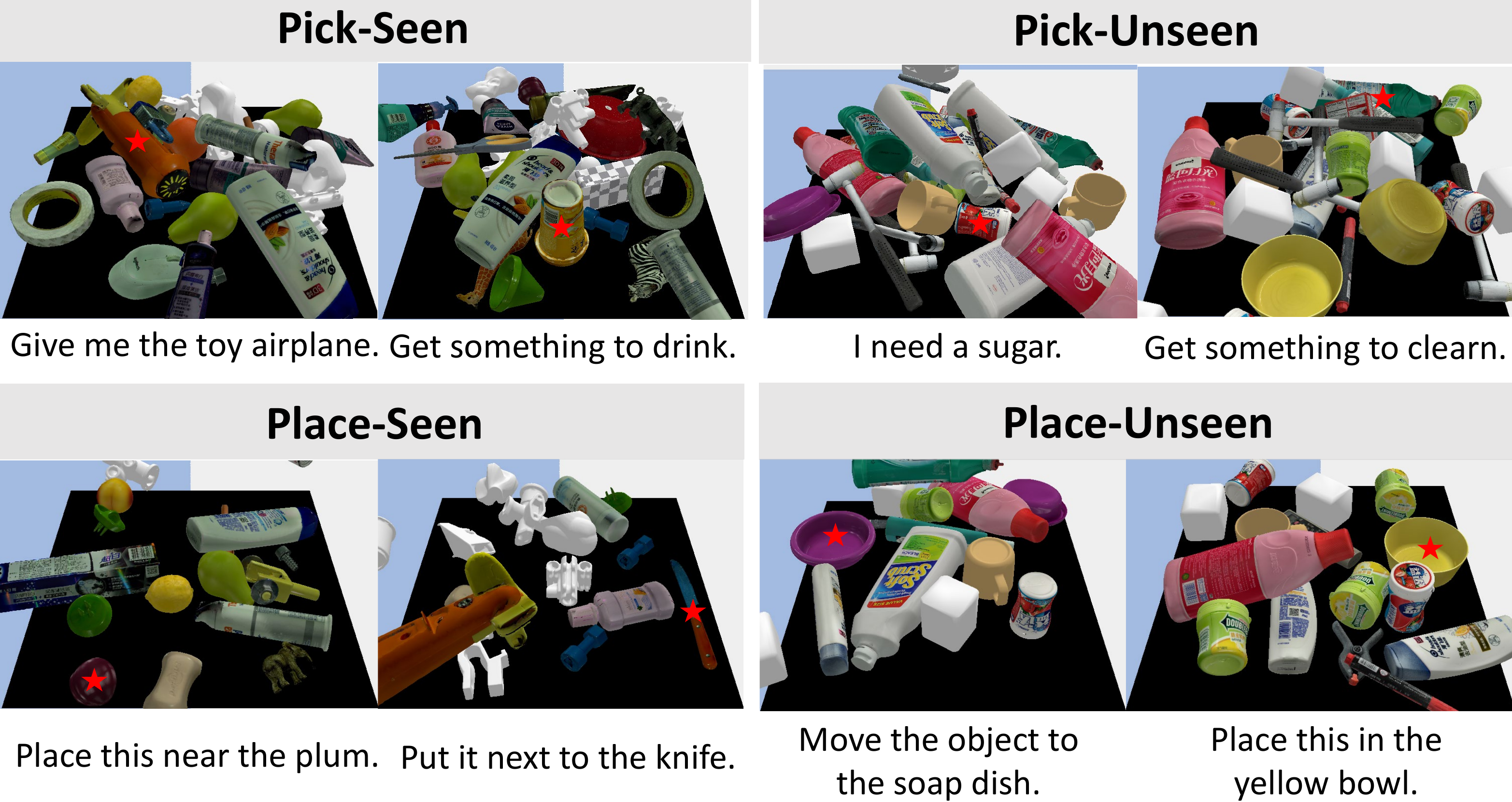}
  % \vspace{-0.35cm}
  \caption{Example test cases of double object number in simulation. The target objects or reference objects are labeled with stars.}
  \label{fig:case_30}
  % \vspace{-0.2cm}
\end{figure}

\subsection{Simulation Experiment Details}

{\bf Test Case Visualizations.} We collect test cases with 66 seen objects and 17 unseen objects. More example test cases across all categories are presented in Fig.~\ref{fig:more_simulation_cases}. 
\revise{
For cases of more objects~(30 objects in a scene) in Sec.~\ref{table:scale}, Fig.~\ref{fig:case_30} shows some example cases, demonstrating more complex settings with frequent occlusion and dense clutter than those in Fig.~\ref{fig:more_simulation_cases}.
}

{\bf Baseline Implementations.} 
For the two neural field based pick policies, we train the feature fields for each step of action planning and select the grasp pose with the maximum query score of language instructions from a given set generated by GraspNet~\cite{fang2020graspnet}. For language queries, we use the object as the positive query ({\it e.g.} ``pear'', ``something to drink''), empty string as the negative query, and ``body'' as the part query.

For LERF-TOGO~\cite{rashid2023lerftogo}, we strictly follow the training and querying codes$\footnote{\href{https://github.com/lerftogo/lerftogo}{https://github.com/lerftogo/lerftogo}}$ provided by the authors. For input data, there are RGB-D images of 53 views including the 3 used by ours and 50 additional different views to provide more visual information, with a format aligned to their example.

For GraspSplats~\cite{ji2024graspsplats},
we used the open-sourced codes$\footnote{\href{https://github.com/jimazeyu/GraspSplats}{https://github.com/jimazeyu/GraspSplats}\label{fn:graspsplats}}$ for static scene grasping, as it is claimed in the paper to achieve better success rates than dynamic scene grasping. The ground-truth poses for each viewpoint are obtained from the simulation, and written directly to the COLMAP~\cite{schoenberger2016sfm} database as required. We run COLMAP for point cloud initialization with ground-truth camera poses. To investigate the influence of image input, we test Graspsplats with 23 view RGB, 3 view RGB-D, and 23 view RGB-D supervision respectively, as shown in Table~\ref{table:supp-graspsplats}. In these experiments, the input images contain the 3 input views used to evaluate our method. It is worth noting that with the default parameters, GraspSplats fails in most of the cases to select grasp poses because its default distance threshold of $0.02m$ excludes most grasp poses generated by GraspNet. To address this, we increased the distance threshold to $0.08m$.

In the original implementation\footref{fn:graspsplats}, GraspSplats first queries for the object, then crops the point cloud using a hard-coded workspace limit. This can result in failures when the queried object lies outside the workspace, leaving the cropped point cloud devoid of the target. To mitigate this issue, in Table~\ref{table:supp-graspsplats} we first crop the point cloud and then query the object, ensuring the target remains within the workspace limit. Therefore, the results of 23 RGB images in Table~\ref{table:supp-graspsplats} are better than others, which avoids some failures of the queried object outside the workspace.

\begin{table}[t]
\caption{Results of GraspSplats with Different Inputs}
\label{table:supp-graspsplats}
\centering
\begin{tabular}{cccc}
\toprule
Data & Seen & Unseen\\
\midrule
\multirow{3}{*}{}
23 RGB & 68.0/1.89 & 31.3/2.19 \\
3 RGB-D & 55.0/3.48 & 34.6/1.36 \\
23 RGB-D & 58.0/2.05 & 37.3/1.667  \\
\bottomrule
\end{tabular}
\begin{tablenotes}
\footnotesize
\centering
\item * Metrics are presented as Task Success Rate / Planning Steps.
\end{tablenotes}
\end{table}

We analyze failure cases of GraspSplats and identify several key issues. A portion of failures stemmed from the collisions with other objects that cause the target object to drop from the gripper. Also, although GraspSplats can correctly segment the queried object, the selected grasp point might be grasping other surrounding objects since objects are closely packed in the clutter. Additionally, GraspSplats is more sensitive to typos in language instructions, which are included in the test cases.

\revise{
\subsection{Analysis of Evaluation Results}
{\bf Unseen Pick Cases.} In Table~\ref{table:baseline}, We observe that unseen objects achieve a slightly higher success rate compared to seen objects under similar planning steps. While this may initially appear counterintuitive, we hypothesize that this phenomenon arises from two complementary factors: the visual distinctiveness and semantic clarity of the target objects relative to other objects in the same scene.

Specifically, we analyze the visual appearances and target object prompts of the unseen cases, and find that most of them are more distinct from surrounding distractor objects~({\it e.g.} less clutter), making them easier for CLIP to differentiate in context. Additionally, their prompts often describe high-frequency, semantically generic categories~({\it e.g.}, box, container). These objects tend to have well-aligned visual-language embeddings in the pretrained CLIP. In contrast, some target objects of seen cases have more ambiguous or less common names~({\it e.g.}, drink, theramed) or more cluttered appearances that make alignment more difficult.

To support this hypothesis, we conduct a CLIP-based similarity analysis. For each target object, we compute the cosine similarity between its point features and the corresponding language prompt~({\it e.g.}, “Give me the box.”). On average, unseen objects achieve higher similarity scores than seen objects (0.30 vs. 0.28), suggesting that the pretrained model is more confident in aligning these objects with the language instruction. We further compute a similarity margin, the difference between the target object's similarity and the mean similarity of other objects in the same scene, as a distinctiveness score. Unseen target objects consistently yield higher distinctiveness scores (0.11 vs. 0.09 on average), indicating that CLIP can distinguish them from distractors more clearly and confidently. Representative examples are shown in Fig.~\ref{fig:case_vis}, where we observe a more concentrated high similarity within the target object in the unseen object case.

To further test our policy in similar condition of seen cases, we collect a set of unseen object cases whose distinctiveness scores are similar to those of seen object cases. Examples are shown in Fig.~\ref{fig:case_unseen}~(b), which demonstrate more visual clutter and semantic complexity. As shown in Table~\ref{table:unseen-case}, these unseen cases yield lower success rates and more planning steps, though performance remains acceptable, further validating our method's generalization ability.

\begin{figure}[t]
  \centering
  % \color{blue}
  \includegraphics[width=0.7\linewidth]{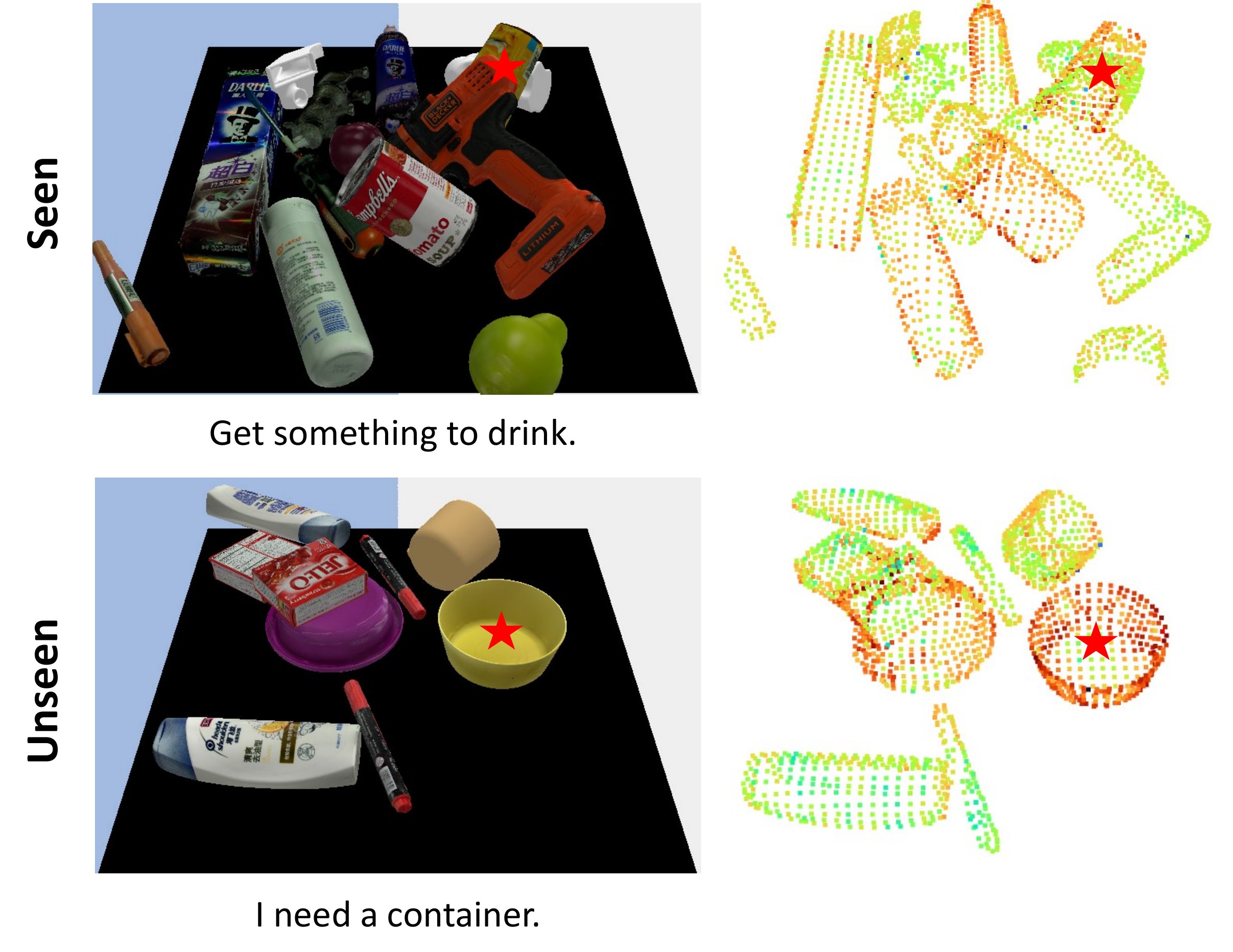}
  % \vspace{-0.35cm}
  \caption{Case visualization of similarity clouds with normalized similarity values. The target objects are marked with red stars. In similarity clouds, the closer the color is to red, the higher the similarity it indicates.}
  \label{fig:case_vis}
  % \vspace{-0.2cm}
\end{figure}

\begin{figure}[t]
  \centering
  % \color{blue}
  \includegraphics[width=\linewidth]{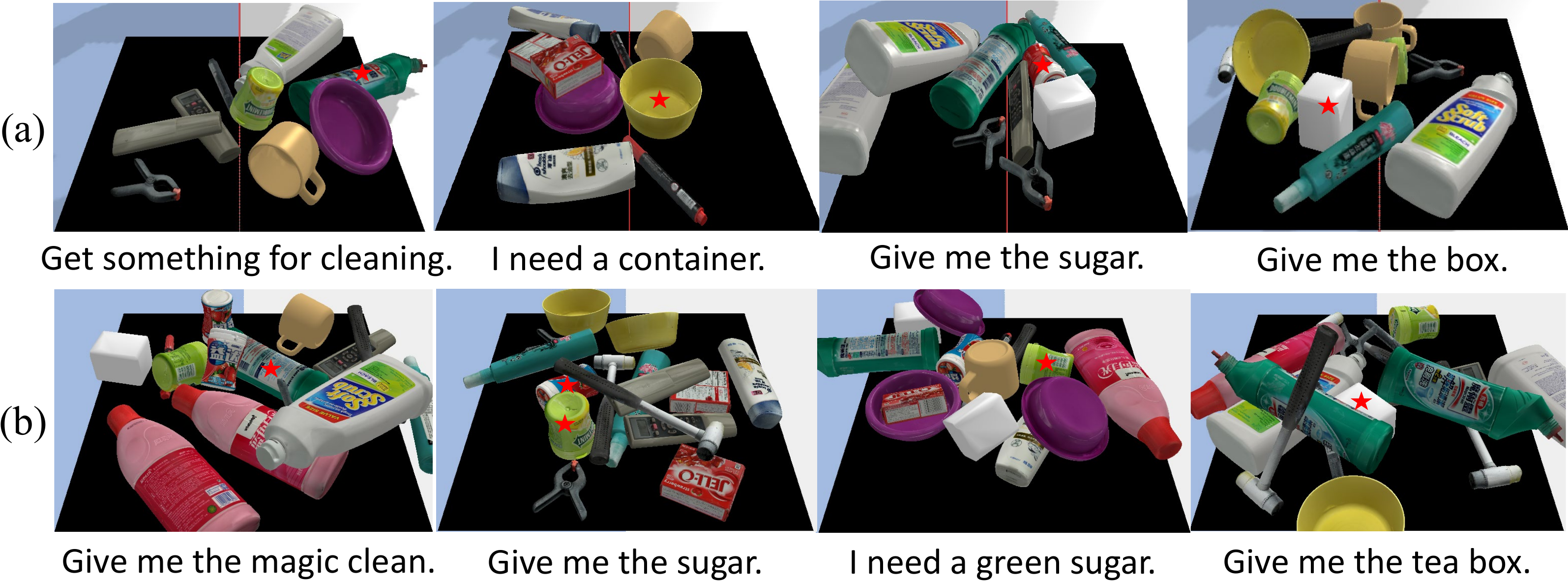}
  % \vspace{-0.35cm}
  \caption{Examples of (a) original unseen pick cases and (b) newly tested unseen pick cases in Table~\ref{table:unseen-case}.}
  \label{fig:case_unseen}
  \vspace{-0.2cm}
\end{figure}

\begin{table}[t]
\centering
% \color{blue}
\caption{Simulation Results on Different Unseen Cases}
\label{table:unseen-case}
\begin{tabular}{cccc}
\toprule
VC $\uparrow$ & SC $\uparrow$ & Task Success & Planning Steps \\
\midrule
& & 97.3 & 2.57 \\
$\checkmark$ & $\checkmark$ & 88.0 & 3.45 \\
\bottomrule
\end{tabular}
\vspace{0.3em}
\begin{tablenotes}
\centering
\footnotesize
\item * VC: visual clutter. SC: semantic complexity. \\
$\uparrow$ represents making the condition harder.
\end{tablenotes}
\vspace{-0.3cm}
\end{table}

{\bf Decomposing Place Task Difficulty.} We observe that place baselines in Table~\ref{table:baseline} demonstrate weak performance. To better understand the performance gap, We conduct an additional set of experiments by decomposing the place task difficulty in language instructions and scenarios. Specifically, we evaluate VLP~\cite{xu2023object} on place tasks under four conditions: (1) simple instruction (SI) with simple scenario (SS), (2) SS only, (3) SI only, and (4) the default setting with flexible instructions and cluttered scenes. For SI, we use phrases and category labels that are more readily by CLIP~({\it e.g.} ``a photo of a yellow cup''). For SS, we test the model on some simple scattered scenarios. The results are summarized in Table~\ref{table:vlp}. Overall, both instruction and scenario simplification significantly improve success rates, particularly in the ``Seen'' setting. The best performance is achieved when both simplifications are applied. These results confirm that the language foundation model (CLIP) struggles with ambiguous instructions and heavy clutter, which limits its standalone performance in complex place tasks. In contrast, by integrating attention-based policy learning with foundation model priors, our method is better equipped to handle flexible language instruction and complex scenarios.

\begin{table}[t]
\centering
% \color{blue}
\caption{Comparison of Different Place Settings for VLP}
\label{table:vlp}
\begin{tabular}{p{0.8cm}p{0.8cm}p{1cm}p{1cm}}
\toprule
SI & SS & Seen & Unseen \\
\midrule
$\checkmark$ & $\checkmark$ & 73.3 & 60.0  \\ 
             & $\checkmark$ & 66.7 & 60.0 \\
$\checkmark$ &              & 50.0 & 30.0 \\
             &              & 40.0 & 20.0 \\
\bottomrule
\end{tabular}
\vspace{0.3em}
\begin{tablenotes}
\footnotesize
\centering
\item * SI: simple instruction. SS: simple scenarios.
\end{tablenotes}
\end{table}

{\bf Role of Residual Block.} In our framework, the multi-modal nature of the policy arises from modeling categorical action distributions, which can naturally represent multiple possible actions for the same input. The role of residual blocks is to facilitate efficient policy adaptation based on a pretrained model, allowing the model to adjust its output distribution using multi-labeled demonstrations while preserving prior knowledge. This technique is widely used in adaptation of vision-language models~({\it e.g.}, CLIP-Adapter~\cite{gao2024clip}), where residual branches help models adapt to new distributions with minimal disruption to pretrained representations. In our case, the residual block shows its strength during fine-tuning on multi-labeled, task-specific data.

{\bf Failure Modes.} Fig.~\ref{fig:case_fail} visualizes some typical failure modes, including heavy occlusion and visual ambiguity of target objects, as well as the semantic ambiguity in language instruction. In the left case, the target object ``strawberry'' is largely occluded by other distractors, demonstrating a heavily cluttered scene. In such cases, the policy struggles to pick up the target within limited planning steps. In the middle case, the target object ``darlix toothpaste'' shares a similar visual appearance with the distractor ``darlix box'', which misleads the policy during selection. The right case illustrates semantic ambiguity in the language instruction. Although the phrase ``into a cylinder'' suggests that the target object is container-like, the expression lacks specificity and may lead the policy to select other cylindrical objects that do not afford containment.

{\bf Real-world Setups.} In our real-world experiments, we initially adopted a single Intel RealSense L515 camera and observed that our policy achieves a good task performance. This result demonstrates that our method generalizes well to limited-view settings, which are common in practical robotic deployments. We also experimented with multi-camera setups but encountered depth interference caused by overlapping structured-light laser patterns. This interference resulted in noisy or unstable depth maps, which affected downstream modules such as GraspNet, whose grasp predictions rely on accurate depth information. As a result, the decision to use a single camera represents a deliberate trade-off between broader observation coverage and depth sensing reliability. 
}

\vfill

\end{document}